\documentclass{article}

\usepackage[preprint]{neurips_2025}

\usepackage{microtype}
\usepackage{graphicx}
\usepackage{tikz}
\usepackage{subfigure}
\usepackage{booktabs} 
\usepackage{xcolor}
\usepackage{amsmath}
\usepackage{amssymb}
\usepackage{mathtools}
\usepackage{amsthm}
\usepackage{multirow}
\usepackage{xspace}
\everypar\expandafter{\the\everypar\looseness=-1}
\usepackage{hyperref}
\usepackage[capitalise]{cleveref}

\theoremstyle{definition}  
\theoremstyle{definition}
\newtheorem{hypothesis}{Hypothesis}
\theoremstyle{plain}
\newtheorem{theorem}{Theorem}[section]
\newtheorem{lemma}[theorem]{Lemma}

\theoremstyle{definition}
\newtheorem{definition}[theorem]{Definition}
\newtheorem{assumption}[theorem]{Assumption}
\theoremstyle{remark}
\newtheorem{remark}[theorem]{Remark}
\newcommand{\dname}{EcoTaskSet\xspace}
\newcommand{\approach}{GREEN\xspace}
\newcommand{\nexps}{1767\xspace}
\DeclareMathOperator*{\argmax}{arg\,max}

\newcommand{\custombox}[2][black]{%
  \tikz[baseline=(char.base)]{
    \node[fill=#1, 
          text=white, 
          inner sep=1pt, 
          outer sep=0pt, 
          minimum size=1em] (char) {#2};
  }%
}

\definecolor{mydarkgreen}{RGB}{0,100,0}
\usepackage[textsize=tiny]{todonotes}

\title{One Search Fits All: Pareto-Optimal Eco-Friendly Model Selection}

\author{
  Filippo Betello\thanks{ 
 Equal Contribution} \\
  DIAG\\
  Sapienza University of Rome\\
  Rome, Italy \\
  \texttt{betello@diag.uniroma1.it} \\
  \And
  Antonio Purificato$^*$ \\
  DIAG\\
  Sapienza University of Rome\\
  Rome, Italy \\
  \texttt{purificato@diag.uniroma1.it} \\
  \AND
  Vittoria Vineis$^*$ \\
  DIAG\\
  Sapienza University of Rome\\
  Rome, Italy \\
  \texttt{vineis@diag.uniroma1.it} \\
  \And
  Gabriele Tolomei \\
  Department of Computer Science \\
  Sapienza University of Rome\\
  Rome, Italy \\
  \texttt{tolomei@di.uniroma1.it} \\
  \And
  Fabrizio Silvestri \\
  DIAG\\
  Sapienza University of Rome\\
  Rome, Italy \\
  \texttt{fsilvestri@diag.uniroma1.it} \\
}

\begin{document}

\maketitle

\begin{abstract}
The environmental impact of Artificial Intelligence (AI) is emerging as a significant global concern, particularly regarding model training.
In this paper, we introduce \textbf{\approach} (Guided Recommendations of Energy-Efficient  Networks), a novel, inference-time approach for recommending Pareto-optimal AI model configurations that optimize validation performance and energy consumption across diverse AI domains and tasks.
Our approach directly addresses the limitations of current eco-efficient neural architecture search methods, which are often restricted to specific architectures or tasks. Central to this work is \textbf{\dname}, a dataset comprising training dynamics from over \textbf{\nexps} experiments across computer vision, natural language processing, and recommendation systems using both widely used and cutting-edge architectures.
Leveraging this dataset and a prediction model, our approach demonstrates effectiveness in selecting the best model configuration based on user preferences. \looseness -1 Experimental results show that our method successfully identifies energy-efficient configurations while ensuring competitive performance.
\end{abstract}

\section{Introduction}
Artificial intelligence (AI) systems, while enabling advancements in numerous fields, come at a substantial computational and environmental cost. Training and inference for large-scale models, including Large Language Models (LLMs), require vast computational resources (e.g. 539 T CO$_2$-eq\footnote{We use the definition of CO$_2$-eq from \href{https://sor.epa.gov/sor_internet/registry/termreg/searchandretrieve/termsandacronyms/search.do?search=\&term=carbon\%20dioxide\%20equivalent&matchCriteria=Contains\&checkedAcronym=true\&checkedTerm=true\&hasDefinitions=false}{Environmental Protection Agency}.} for the LLAMA 2 model \citep{touvron2023llama2openfoundation}), resulting in consider carbon emissions and raising urgent concerns amid global efforts to combat climate change \citep{10.1145/3442188.3445922,faiz2024llmcarbon}. 
While some models, such as DeepSeek \citep{deepseek}, have attempted to employ new structures and more efficient resource utilization, the prevailing trend continues towards increasingly large and complex models. This reliance on scale worsens the issue, as the drive for performance often ignores its environmental costs \citep{wu2022sustainable, george2023environmental}.

Nonetheless, while increasing attention is given to the environmental impact produced in the training and deployment phase, the energy costs of AI actually begin earlier, at the model selection and optimization stage. This phase, often underreported, involves extensive experimentation to identify the optimal model configuration\footnote{Throughout this paper, we refer to \textit{model configuration} as a specific combination of neural architecture model and training-related parameters, namely batch size and learning rate.}, contributing to a significant share of the overall energy footprint \citep{from_clicks_to_carbon}. 
Developing methods that can predict energy-efficient configurations \textit{before} training begins would, therefore, not only reduce emissions and computational overhead but also shorten the model selection process. On top of that, at a higher technical level, current approaches to eco-efficient Neural Architecture Search (NAS) methods still face the same challenges as traditional NAS: they are computationally expensive \citep{strubell2020energy} and often tailored to specific datasets or architectures, limiting their generalization to diverse tasks and domains \citep{liu2022survey}.

Recent efforts have focused on mitigating this impact by optimizing hardware usage \citep{chung2024reducing, you2023zeus} and reducing the search space \citep{guo2020breaking}. For example, EC-NAS \citep{ec_nas} extends this by optimizing both accuracy and energy consumption for image classification, but it is limited to predefined layer types. CE-NAS \citep{ce_nas}, 
leverages reinforcement learning to optimize NAS algorithms based on GPU availability, but similarly restricts the search to a narrow set of layer types.
To support these efforts, benchmarks like NAS-Bench-101 \citep{ying2019bench} have been proposed to enable energy-focused NAS evaluations.

Given these constraints, it would be highly beneficial to predict a model's performance in terms of accuracy and energy consumption before execution. For instance, consider a scenario where a large-scale Neural Network (NN) for image classification requires dozens of experiments to fine-tune the number of layers, learning rate, and regularization methods. Predicting an optimal configuration upfront could eliminate the need for extensive trial-and-error runs, saving hundreds of GPU hours and avoiding significant CO$_2$-eq emissions.

This paper introduces a novel method named \textbf{\approach} (Guided Recommendations of Energy-Efficient Networks) recommending Pareto-optimal NN configurations that balances expected performance on a validation set and energy consumption for any dataset and task across three distinct AI domains, namely computer vision, natural language processing (NLP) and recommendation systems. Crucially, this process operates entirely at inference time.
Unlike existing approaches in energy-efficient multi-objective NAS, our method is highly flexible and extensible across multiple domains. It can be extended to any number and type of objectives, architectures, and datasets in the aforementioned domains, making it suitable for diverse applications. From an implementation standpoint, our approach leverages a custom multi-domain knowledge base, \textbf{\dname}, constructed from over \textbf{\nexps} NN training processes.

\begin{figure*}
    \centering
    \includegraphics[width=\linewidth]{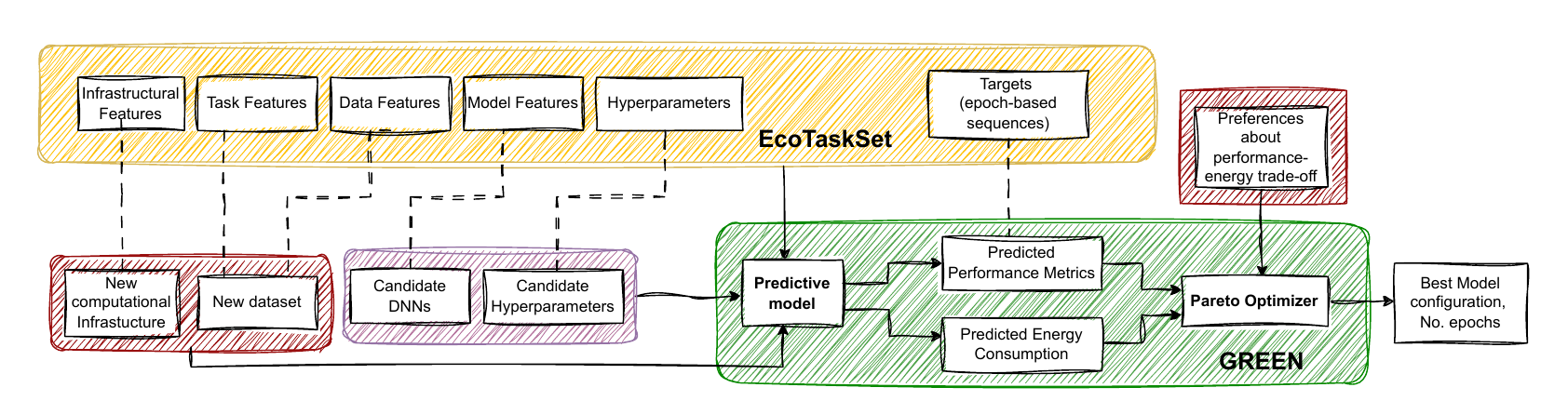}
    \caption{An overview of \approach. It takes as input features from \dname. Then \approach identifies energy-efficient configurations while maintaining competitive performance metrics. The output is a set of Pareto-optimal model configurations, which can be ranked according to user preferences to suggest the \textit{single} best model configuration for a specific dataset, task, and computational infrastructure.}
    \label{fig:graphical_abstract}
\end{figure*}

Overall, the main contributions of our work can be summarized as follows:

\textbf{(1)} We introduce \textbf{\approach}\footnote{The anonymous code is available \href{https://anonymous.4open.science/r/carbon_best-81DE/}{here}.}, a new method to provide multi-objective Pareto-optimal solutions for selecting the best model configurations completely at inference time and differs from current literature by being extensible to any number and type of objectives, architectures, and datasets. The overall approach is depicted in \cref{fig:graphical_abstract}. 

\textbf{(2)} We create and release to the community \textbf{\dname}\footnote{We release the anonimised dataset \href{https://drive.google.com/drive/folders/1lXdSsW2FRU331bpGWOsXrcg-Bp3Px4Pi?usp=sharing}{here}.}, a dataset capturing neural network training dynamics across three domains: computer vision, natural language processing, and recommendation systems. It includes both well-established and cutting-edge, ready-to-use neural architectures that are widely adopted in real-world application scenarios. Moreover, unlike existing benchmarks, it provides detailed epoch-level metrics on both validation performance and energy consumption, offering a valuable resource for research in eco-efficient machine learning and the study of deep learning training dynamics.

\textbf{(3)} We introduce \textbf{SOVA} (Set-Based Order Value Alignment), a new ranking alignment metric designed to evaluate the alignment of true multi-objective metric values across two ranked sets.

\textbf{(4)} Extensive experiments demonstrate  that \textbf{\approach} successfully identifies energy-efficient configurations while maintaining competitive performance metrics.


\section{Related Work}
\label{sec:related}
Widely used NAS algorithms like DARTS \citep{liu2018darts} and Efficient NAS \citep{elsken2018efficient}, are known for being highly CO$_2$-intensive \citep{strubell2020energy}. Recent studies have explored ways to mitigate this environmental impact, either by optimizing hardware usage \citep{chung2024reducing, you2023zeus} or by reducing the search space to make NAS more efficient \citep{guo2020breaking}.
In parallel, several benchmarks \citep{ea_has_bench,ec_nas,wang2020benchmarking} and frameworks for efficient NAS have been introduced \citep{ce_nas}. 

\paragraph{Exisiting benchmarks.}
A wide range of benchmark datasets have been created to facilitate research in architecture search and efficiency-aware learning. However, these datasets often operate under restrictive design choices that limit their applicability to real-world scenarios.
The NAS-Bench family of datasets—NAS-Bench-101 \citep{ying2019bench}, NAS-Bench-201 \citep{dong2020bench}, and NAS-Bench-301 \citep{zela2020surrogate} define fixed, low-complexity search spaces over small-scale convolutional architectures, typically on datasets such as CIFAR-10 and CIFAR-100. NAS-Bench-101 supports only three operation types and a constrained graph topology. NAS-Bench-201 marginally expands the space to support different datasets but still excludes transformers or other contemporary architectures. NAS-Bench-301 introduces a larger search space, but relies on surrogate models trained on partial data, introducing approximation artifacts that reduce reliability, especially under distribution shifts. Crucially, none of these benchmarks provide per-epoch energy measurements.
General-purpose performance prediction benchmarks such as LCBench \citep{ZimLin2021a} and Taskset \citep{metz2020using} offer broader task coverage but similarly abstract away the training process. LCBench focuses on tabular datasets and shallow MLP architectures, providing only scalar performance metrics and metadata under fixed hyperparameters. Taskset focuses exclusively on RNN models and NLP tasks. Importantly, neither benchmark includes energy consumption tracking or system-level resource information.\\
These benchmarks, while useful within their respective domains, fall short of supporting sustainability-focused research or enabling fine-grained study of training behavior across architectures and domains.

\paragraph{Eco-Aware NAS methods.} In parallel, recent methods have extended NAS algorithms to incorporate energy or hardware-awareness.
For instance, \cite{ec_nas} proposed EC-NAS a benchmark focused on energy-aware NAS for image classification, built upon the foundational NAS-Bench-101 dataset \citep{ying2019bench}. EC-NAS enables multi-objective NAS by identifying models that balance energy consumption and accuracy. It outputs a Pareto frontier of optimal models, but restricts architectural choices to a predefined set of layers (e.g., 3x3 convolution, 1x1 convolution, 3x3 max pooling). However, EC-NAS has notable limitations: it reports performance metrics only at a few predefined epochs, inherits the restricted architectural diversity of NAS-Bench-101, and assumes a fixed threshold budget during search, which limits flexibility in exploring trade-offs between energy and performance.
\cite{ce_nas} introduced CE-NAS, a framework that uses \citep{dong2020bench,siems2020bench}, which optimizes NAS architecture selection based on GPU availability. They proposed a reinforcement learning-based policy to allocate NAS algorithms across clusters with multiple GPUs. However also CE-NAS restricts its search space to a small set of layer types. \cite{xu2021knas}proposed KNAS, a gradient-based method that is able to evaluate randomly-initialized networks. It achieves large speed-up in NAS-Bench-201 benchmarks \citep{dong2020bench}; however, similar to the previously discussed approaches, also KNAS is constrained by the limited architectural diversity provided by the underlying benchmark. 

To address these gaps, we present \dname, a benchmark dataset, and \approach, a method for jointly recommending energy-efficient configurations—spanning neural architecture, training budget, and key hyperparameters—based on realistic, domain-diverse training runs.

\section{\approach}
\label{sec:approach}
\looseness -1   In this study, given task and dataset, we address the problem of selecting the optimal model configuration  while considering user's preferences regarding the trade-off between performance and environmental impact.
From a practical viewpoint, we claim that the task of identifying the optimal \textit{model configuration} for a given machine learning problem can be approached as a combined learning and optimization challenge. For this reason, we propose a solution based on a two-step approach. The first step involves a learning and prediction phase, which leverages a cross-domain knowledge base (\dname). The second step encompasses a multi-objective optimization and preference-based ranking to select the optimal model configurations, given task and dataset. Owing to space limitations, the computational complexity analysis of the algorithm is deferred to \cref{app:additional_studies}, while the remainder of this section focuses on the theoretical foundations and formalization of the proposed solution.

\subsection{Theoretical Foundations}
\label{sec:theor_found}
\begin{assumption}[Non-linear Relationship between Performance and Energy]
\label{ass:nonlinear}
  Let $I$ be a given computational infrastructure.
  For a neural network model $M$ trained on task $T$ with dataset $D$,
  we denote by $A_e$ the performance on the validation set 
  and by $E_e$ the energy consumption at epoch $e$.
  We assume there exist  a non-linear function: 
  \[
    A_e, E_e = f(\phi_T, \phi_D, \phi_M, \phi_I, \theta, e),
  \]
  where 
  \(\phi_T\) describes the task, 
  \(\phi_D\) the dataset, 
  \(\phi_M\) the model configuration,
  \(\phi_I\) the infrastructure characteristics, 
  and \(\theta\) training hyperparameters.
  This expresses that both $A_e$ and $E_e$ depend on these features in a complex, non-linear way.
\end{assumption}

\begin{hypothesis}[Sufficiency of Feature Descriptors]
\label{prop:sufficiency}
  Under Assumption~\ref{ass:nonlinear}, 
  a sufficiently rich set of descriptive features 
  (\(\phi_T, \phi_D, \phi_M\, \phi_I\)) 
  allows us to approximate, with small error, 
  the variation in energy consumption and validation performance across different epochs and configurations. 
\end{hypothesis}

\begin{hypothesis}[Neural Network as Universal Approximator]
\label{hyp:nn_approx}
  We posit that the function $f$ in Assumption~\ref{ass:nonlinear} 
  can be effectively approximated by a neural network, 
  leveraging cross-domain knowledge and the interplay among the various features. 
  Furthermore, such a neural network is capable of generalizing 
  to new epochs and novel model configurations,
  thus providing a flexible framework for predicting energy consumption and performance.
\end{hypothesis}

\begin{remark}
We emphasize that $E$ is our main focus because carbon intensity, which is used to estimate carbon emissions,
is determined by the energy mix specific to the geographical location where the computation occurs and it represents a multiplicative factor with respect to $E$.
As shown in \cite{faiz2024llmcarbon}, in fact, the carbon footprint of a computational process directly correlates 
with the carbon intensity of the electricity used.
However, since the carbon intensity is independent of the energy consumed in a computational process, we prefer focusing directy on $E$.
This choice allows us to align more directly with the existing literature on energy-efficient computation, 
  placing our work within a broad line of research 
  aimed at reducing the overall energy footprint of machine learning.
\end{remark}

\subsection{Inputs}
\label{sec:features}
Given what we assume and claim in \cref{sec:theor_found}, we define the input space $ \mathcal{X} =\{{\mathcal{M}, \mathcal{T}, \mathcal{D}, \mathcal{I}} \}$, where
$\mathcal{M}$ denotes a set of NN model configurations,
$\mathcal{T}$ a set of tasks, $\mathcal{D}$ a set of datasets, and $\mathcal{I}$ a set of computational infrastructures. This representation allows any machine learning problem to be expressed as a tuple $(M, T, D, I) \in \mathcal{M} \times \mathcal{T} \times \mathcal{D} \times \mathcal{I}$, 
where each combination encapsulates the interactions between the model, task, dataset, and computational environment, collectively influencing system performance and resource consumption.
For each set $\mathcal{X}_k  \in \mathcal{X}$, we define:  
\[
\mathcal{X}_k = \{ X_{k,i} \}_{i=1}^{|\mathcal{X}_k|} \quad \text{and} \quad  \phi(X_{k,i}) = \{ x_{k,i}^j \}_{j=1}^{|\phi(X_{k,i})|}
\]
where each element $X_{k,i}$ is represented by a feature vector $\phi(X_{k,i}) \in \Phi_{\mathcal{X}_k}$, with $\Phi_{\mathcal{X}_k}$ denoting the feature space associated with the set $\mathcal{X}_k$. Notably, these feature spaces are are theoretically unbounded, allowing for infinite variability in configurations, domains, and data sources. Each feature $x_{k,i}^j$ corresponds then to a measurable property of $X_{k,i}$.  
\looseness -1 It should be noticed, that despite being formally associated with a specific set for modeling reasons, some features, from a practical standpoint, span multiple dimensions. For instance, the number of floating-point operations for a model $M$ depends on both its architecture and the dataset characteristics.

The detailed design of these feature spaces and their associated feature sets are  in \cref{appendix:kb}.

\subsection{Predictive Model Learning }
In the first step of our approach, we aim to construct a predictive function $ q_\theta $ that approximates the function $ f $ introduced in  \cref{sec:theor_found}, which is able to estimate the validation performance $ A $ and energy consumption $ E $ of a model $ M$ at epoch $ e $, for a given task $ T $, dataset $ D $ and computational infrastructure $ I $. Since \cref{ass:nonlinear} models $ A $ and $ E $ as non-linear functions of task features ($ \phi_T $), dataset features ($ \phi_D $), model configuration features ($ \phi_M $), training hyperparameters ($ \theta $), the epoch ($ e $), and the computational infrastructure ($ I $), formally, we define:
\[
q_\theta: \mathcal{X} \to \mathcal{Y}, \quad 
\mathcal{X} = \Phi_T \times \Phi_D \times \Phi_M \times \Theta \times \mathbb{N} \times \mathcal{I}, \quad
\mathcal{Y} = \mathbb{R} \times \mathbb{R}_{\geq 0}.
\]

such that $q_\theta(\phi_T, \phi_D, \phi_M, \phi_I,\theta, e ) = ({A}_e, {E}_e )$ where $ \Theta $ is the space of training hyperparameters for $q_\theta$ and $ \mathbb{N} $ is the epoch space. Here, $ A \in \mathbb{R} $ is a task-dependent performance metric (e.g., accuracy for classification tasks or mean squared error for regression tasks) and $ E \in \mathbb{R}_{\geq 0} $ represents an environmental impact metric, such as, in our case, the energy consumption.
To simplify notation, we henceforth write $ q_\theta(\Phi, \theta, e) $, where 
$\Phi = \Phi_T \cup \Phi_D \cup \Phi_M \cup \Phi_I$.

To determine the parameters $ \theta \in \Theta $, we minimize a step-wise weighted loss function that balances the prediction of performance ($A$) and energy consumption ($E$) over a sequence of training epochs. The optimization objective is:
\[
\theta^* = \arg\min_{\theta \in \Theta} \mathbb{E}_{(\Phi, \theta, e) \sim p} \left[ \mathcal{L}\bigl(q_\theta(\Phi, \theta, e), (A_e, E_e), \alpha_e\bigr) \right],
\]

where $ \mathcal{L} $ is the composite loss function for a given epoch $ e $ and uses the Mean Absolute Error (MAE) as the base metric and $\alpha_e \in [0,1]$  is the weight of the energy-related loss component. For predicted values $ \hat{A}_e $ and $ \hat{E}_e $, and true values $ A_e $ and $ E_e $, the step-wise MAE losses are computed as:
\[
\mathcal{L}_{A, e} = \frac{1}{B} \sum_{i=1}^B |\hat{A}_e^{(i)} - A_e^{(i)}|, \quad
\mathcal{L}_{E, e} = \frac{1}{B} \sum_{i=1}^B |\hat{E}_e^{(i)} - E_e^{(i)}|,
\]

where $ B $ is the batch size, and $ e \in \{ 1, \ldots, V \}$, with $ V $ being the maximum number of epochs. The composite loss at each epoch is given by:
\[
\mathcal{L}_{\text{comp}, e} = \alpha_e \mathcal{L}_{A, e} + (1 - \alpha_e) \mathcal{L}_{E, e},
\]

where the dynamic weights $\alpha_e$ are computed based on the relative rates of change of the individual losses. First, we calculate the rates of change and normalize them to compute the weight $\alpha_e$ for the loss $\mathcal{L}_{A, e}$ such that:
\[
r_{A, e} = \frac{\mathcal{L}_{A, e}}{\mathcal{L}_{A, e-1}}, \quad 
r_{E, e} = \frac{\mathcal{L}_{E, e}}{\mathcal{L}_{E, e-1}}, \quad \alpha_e = \frac{r_{A, e}}{r_{A, e} + r_{E, e}}.
\]

For initial epochs ($ e < 2 $), where sufficient historical data is unavailable, equal weights are assigned: $ \alpha_e = 0.5 $. The overall loss for the training process is then computed as the average of the composite losses over all epochs:
\[
\mathcal{L} = \frac{1}{V} \sum_{e=1}^V \mathcal{L}_{\text{comp}, e}.
\]

\subsection{Multi-Objective Optimization and Ranking for Best Model Selection}

Once $ q_\theta $ has been learned, the next step is to identify the optimal model configuration denoted as $ (M^*, e^*) $, that satisfies user-defined preferences for the trade-off  between performance ($\omega_A$) and energy consumption ($\omega_E$). 

A naive strategy involves identifying the model configuration and epoch that minimizes the predicted energy consumption $ \hat{E} $, subject to a user-set performance constraint ($\hat{A} \geq \gamma $), where $ \gamma $ is the minimum performance level required by the user. Although effective in optimizing energy consumption while meeting a fixed performance threshold, such approaches inherently prioritize one objective over the other and fail to account for the trade-offs between performance and energy consumption.
\looseness -1 To address this limitation, we formulate the task as a multi-objective optimization problem, aiming to simultaneously maximize $ \hat{A} $ and minimize $ \hat{E} $. The optimization proceeds in two stages reported below.

\subsubsection{Pareto Frontier Identification}
In the first stage, we compute the Pareto frontier \citep{pareto1964cours}, which identifies all non-dominated solutions where no other configuration achieves better performance with lower energy consumption. Mathematically, a solution $(M_i, e_j)$ is Pareto-optimal if there exists no other solution $(M_{i'}, e_{j'})$ s.t.:
\[
\hat{A}(M_{i'}, e_{j'}) \geq \hat{A}(M_{i}, e_{j}) , \quad \hat{E}(M_{i'}, e_{j'})\leq \hat{E}(M_{i}, e_{j}) ,
\]
with at least one strict inequality. Constructing the Pareto frontier $\mathcal{P}$ allows to reduce the search space to configurations that represent the best trade-offs between $ \hat{A} $ and $ \hat{E} $. The procedure we used to identify the Pareto frontier is presented in \cref{app:additional_studies}.

\subsubsection{Preference-Based Filtering and Ranking}
In the second stage, we employ a preference-based filtering and ranking method to select either multiple or single solutions from the Pareto frontier $\mathcal{P}$, based on user-defined preferences. This approach enables tailored decision-making by allowing users to define specific criteria for selection. For instance, solutions can be filtered based on a minimum performance threshold $\gamma$, ensuring that only configurations meeting user-specified baseline requirements are considered.
If a single solution must be selected, various ranking methods can be applied to capture different prioritization strategies, such as proximity to optimal outcomes (e.g., distance to the ideal point) or user-defined preferences regarding the relative importance of one metric over another. In this context, user-defined weights $(\omega_{A}, \omega_{E})$, where $\omega_{A} + \omega_{E} = 1$, can be used to represent the trade-off between validation performance and energy consumption. The score for a given configuration is then defined as:
\[
S(M, e) = \omega_{A} \hat{A}_e - (1 - \omega_{E}) \hat{E}_e.
\]
In this case the optimal solution is then:
\[
(M^*, e^*) = \argmax_{(M, e) \in \mathcal{P}} S(M, e).
\]

\subsection{Online Updates}

Since predictive accuracy of $ q_\theta $ is critical for robust recommendations, in a real-world scenario, $ q_\theta $ must adapt to evolving task, dataset, and model spaces to remain effective.
To achieve this, the parameters $ \theta $ can be refined in an online learning fashion. For the selected model configuration $ M_{e}^*=(M^*, e^*) $ actually trained on dataset $D{_i}$ to solve task $T{_i}$ with computational infrastructure $I_i$ the update rule is:
\[
\theta \leftarrow \theta - \eta \sum_{\tilde{e}=0}^{e^*} \nabla_{\theta} \mathcal{L}\left(q_{\theta}(M_{\tilde{e}}^*), (P(M_{\tilde{e}}^*) , E(M_{\tilde{e}}^*)),\alpha_{\tilde{e}}\right),
\]

\noindent
where $q_{\theta}(M_{\tilde{e}}^*)=q_{\theta}(\phi_{T_i}, \phi_{D_i}, \phi_{M{^*}},\phi_{I_i}, \theta, \tilde{e} )$, $ \eta $ is the learning rate, $ \tilde{e} \in [0, e^*] $ spans epochs from the initial one to $ e^* $. Lastly  $P(M^*, \tilde{e})$ and  , $E(M^*, \tilde{e})$ are the  performance and energy metrics obtained during the actual training with the suggested model configuration.

\section{\dname}
\label{sec:dataset}
 Unlike prior benchmarks that operate in synthetic or narrow domains, \dname is built from diversified training runs across three major areas of machine learning practice: computer vision, natural language processing, and recommendation systems. For each run, we log per-epoch validation accuracy, energy consumption, and system-level details.
The models, datasets and the tasks used to create this knowledge base (KB) can be found in \cref{tab:knowledge_base_summary} and they are described in details in \cref{appendix:d_m_details}. 
Selected well-known and established model architectures are trained for a domain-dependent number of epochs, using three different learning rates and five batch size values to account for variability in optimization dynamics.
To also investigate the influence of dataset size on training dynamics, we remove varying percentages of samples from the data, ensuring an equal proportion from each class. To track the energy consumption of all the experiments we use CodeCarbon 
 \citep{codecarbon}, a tool designed to track the power consumption of both CPUs and GPUs as well as additional metrics like CO$_2$-eq and total energy consumed. From all the samples, we extract key information that forms the features in our dataset, as described in Section \ref{sec:features}: hyperparameters, infrastructural features, task features, data features and model features.

Comprehensive details of the model configurations used to build the KB and the features of our dataset are provided in \cref{appendix:kb}.
Each sample from the dataset has two important features, which we consider the targets for \approach: the validation metric at the selected epoch (i.e., accuracy for image classification or F1-score for text classification) and then the energy emission at the same epoch, computed via CodeCarbon. The number of samples in the dataset is \nexps. 

\begin{table*}[ht]
\resizebox{\textwidth}{!}{
\begin{tabular}{cccc}
\toprule 
Domain & Task & Dataset & Model \\
\midrule
Computer Vision & Classification &  FOOD101, MNIST, Fashion-MNIST, \underline{CIFAR-10} & AlexNet, EfficientNet, ResNet18, SqueezeNet, ViT, VGG16 \\
NLP & Q\&A, Sentiment Analysis & Google-boolq, StanfordNLP-IMDB, Dair-ai/Emotions, \underline{Rotten\_tomatoes} & RoBERTa, BERT, Microsoft-PHI-2, Mistral-7B \\ 
Recommendation Systems & Sequential Recommendation & FS-NYC, ML-100k, ML-1M, \underline{FS-TKY} & Bert4Rec, GRU4Rec, CORE, SASRec \\
\bottomrule
\end{tabular}}
\caption{Overview of the Knowledge Base used to train $ q_\theta $. \dname was created selecting 3 different domains, for a total of \nexps experiments. The \underline{underlined} datasets are used for testing. A detailed description of datasets and models can be found in \cref{appendix:d_m_details}.}
\label{tab:knowledge_base_summary}
\end{table*}

\section{Experiments}
\label{sec:exps}
\subsection{Experimental Setup}
We implement the predictive function  $q_\theta$ as a transformer-based NN with 4 transformer encoder layers, each with a dimension of 256 and 8 attention heads. The feed-forward network within each encoder layer has a dimension of 512. The network is specifically designed for multivariate time series inputs, providing multi-target predictions. The hyperparameter configuration was derived through a systematic HPO process, specifically designed to minimize the MAE between the predicted metrics and the ground truth values. 
To validate the correctness of our approach, we conduct experiments on different machines, performing multiple runs and reporting the average results. Details about the hardware configurations and hyperparameter optimization (HPO) settings are provided in \cref{app:hardware_spec}.

We use CIFAR-10, Foursquare-TKY (hereafter, FS-TKY), and Rotten\_tomatoes datasets for testing, while the others are used for training our predictor model (Table \ref{tab:knowledge_base_summary}). 
In our testing setup, we evaluate the results from three independent training runs of the aforementioned predictor model, each initialized with a different random seed. 
The minimum threshold for validation accuracy used to filter the data and construct the Pareto fronts is set to $0.9$ for CIFAR-10 and FS-TKY, and $0.45$ for Rotten\_tomatoes. The lower threshold for Rotten\_tomatoes is due to the limited training dynamics tracked in the NLP experiments, which cover only 5 epochs, resulting in model configurations that, on average, achieve lower validation performance. We compare our approach against different baselines, further described in \cref{app:comp_det}.

\subsection{Evaluation Metrics}
The evaluation of our approach is twofold: $(i)$ assessing the accuracy of the predictor model and $(ii)$ evaluating the alignment between the predicted Pareto front and the true Pareto front (i.e., based on the ground truth values of the target metrics). 
First, we evaluate the accuracy of the learned function $q_\theta$ in predicting the two target metrics: \textit{validation accuracy} and \textit{cumulative energy consumption}\footnote{The cumulative energy target is normalized to the range $[0, 1]$, similar to the validation accuracy, to ensure comparability and stability across different datasets and tasks.}. We assess these predictions at each training epoch using MAE.
Second, we assess the alignment between the predicted Pareto fronts, \( \mathcal{P}_{\text{pred}} \), and the true Pareto fronts, \( \mathcal{P}_{\text{true}} \), using two of the most widely adopted metrics for evaluating solution sets in multi-objective optimization \citep{li2019quality}, specifically the Hausdorff distance (HaD) \citep{henrikson1999completeness, 6151115} and the Hypervolume difference ($\Delta HV$) \citep{zitzler1998multiobjective}. While HaD measures the maximum distance between the nearest points in the two sets, providing a robust indication of how closely the predicted front approximates the true front, $\Delta HV$ captures the difference in the dominated space, offering insight into the extent to which the predicted Pareto front covers the true one.
Additionally, we employ standard classification metrics, such as Recall and F1-score, to assess the effectiveness of our approach in identifying relevant solutions. Lastly, since our ultimate goal is to recommend the optimal model configuration based on the problem setup and user preferences, we evaluate the accuracy of ranking configurations using the Normalized Discounted Cumulative Gain (NDCG) \citep{wang2013theoretical}. Due to space constraints, we define all aforementioned metrics in \cref{app:metrics}.

However, while these metrics effectively measure distance between the Pareto fronts and the ranking consistency, they do not directly account for alignment between ranked Pareto-optimal solutions. To address this limitation, we introduce \textbf{Set-Based Order Value Alignment at k (SOVA@k)}, a ranking alignment metric specifically designed for multi-objective evaluation. Unlike traditional metrics, SOVA@k allows for the comparison of two sets—such as a true Pareto front and a predicted Pareto front—that may contain different items, provided both sets have the same number of ranked elements $(k)$ \footnote{In scenarios where the true and predicted Pareto fronts have the same number of ranked elements ($k$), SOVA@k effectively compares these sets. However, when ties occur, multiple items may share the same rank position, leading to sets of different lengths. To address this, SOVA@k can be extended to handle sets of varying lengths by appropriately adjusting the ranking positions to account for tied items. For a detailed explanation of this extension, please refer to the \cref{app:sova_expanded}.}. Notably, it evaluates the alignment between two ranked sets not based solely on the order of items, but on the true values of relevant metrics. Specifically, it quantifies ranking alignment by computing the weighted sum of absolute differences in true objective values, applying rank-based weighting to prioritize higher-ranked positions and user-defined objective weighting to emphasize the relative importance of different objectives:
\begin{definition}  
    Given a set of items \( I \), where each item is characterized by \( m \) objectives, and two ranked Pareto frontiers \( X = (x_1,...,x_k) \) and \( Y = (y_1,...,y_k) \), where \( x_i, y_i \in I \), with \( k \in \mathbb{N^+} \). Let \( w_i \) be position weights and \( \tau_j' \) be the normalized objective weights. The Set-Based Order Value Alignment (SOVA) at \( k \) is defined as:
    \begin{equation*}
        \boldsymbol{\mathrm{SOVA(X, Y)@k}} = {\sum_{i=1}^{k} w_i \cdot \sum_{j=1}^{m} \tau_j' \cdot | {x}_{ij} - {y}_{ij} |},
    \end{equation*}
\end{definition}
where \( k \) is the number of top-ranked elements, \( m \) is the number of objectives, and \( x_{ij} \), \( y_{ij} \) denote the normalized true values of the \( j \)-the objective at rank \( i \) in the true and predicted sets, respectively. Two sets are ranked independently before being passed to the function, and \( w_i \) is the rank-based weight for position \( i \), calculated using exponential decay, while the user-defined objective weight \( \tau_j \) is normalized:
    \[
    w_i = \frac{e^{-\lambda i}}{\sum_{l=1}^{k} e^{-\lambda l}} \quad , \quad  \tau'_j = \frac{\tau_j}{\sum_{l=1}^{m} \tau_l},
    \]
where \( \lambda\! > \!0 \) controls the decay rate, ensuring \( \sum_{i=1}^{k} w_i \!= \!1 \). 

SOVA@k ranges in \([0,1]\), where a value of 0 indicates perfect alignment between the two sets, and a value of 1 signifies complete dissimilarity between them. Proofs of boundedness, additional details of this metric, and a comparison with other metrics can be found in \cref{app:sova_metric}.

\begin{table*}[ht]
    \centering
    \resizebox{\linewidth}{!}{
    \begin{tabular}{ccccccc}
    \toprule
    Dataset & MAE$_{\text{A}}^{0}$ ($\downarrow$) & MAE$_{\text{E}}^{0}$ ($\downarrow$) & MAE$_{\text{A}}^{30}$ ($\downarrow$) & MAE$_{\text{E}}^{30}$ ($\downarrow$) & MAE$_{\text{A}}^{70}$ ($\downarrow$) & MAE$_{\text{E}}^{70}$ ($\downarrow$) \\
    \midrule
    CIFAR-10 & 0.116 $\pm$ 0.002 & 0.014 $\pm$ 0.001 & 0.122 $\pm$ 0.003 & 0.010 $\pm$ 0.000 & 0.163 $\pm$ 0.010 & 0.012 $\pm$ 0.004 \\
    FS-TKY & 0.025 $\pm$ 0.001 & 0.034 $\pm$ 0.001 & 0.024 $\pm$ 0.002 & 0.029 $\pm$ 0.001 & 0.021 $\pm$ 0.002 & 0.030 $\pm$ 0.001 \\
    Rotten\_tomatoes & 0.123 $\pm$ 0.022 & 0.014 $\pm$ 0.000 & 0.140 $\pm$ 0.018 & 0.019 $\pm$ 0.003 & 0.128 $\pm$ 0.024 & 0.035 $\pm$ 0.003 \\
    \bottomrule
    \end{tabular} }
    \caption{Mean $\pm$ Std of MAE between the predicted values (A for validation accuracy, E for energy) from the \approach predictor model and the corresponding ground truth values. The superscript {0, 30, 70} indicates the percentage of samples discarded in each experiment.}
    \label{tab:mae_performance}
\end{table*}

\begin{table}[ht]
    \centering
    \begin{minipage}{0.4\textwidth}
        \centering
         \resizebox{\textwidth}{!}{
        \begin{tabular}{cc}
        \toprule
             Dataset & NDCG ($\uparrow$)  \\
        \midrule
             CIFAR-10 & 0.985 $\pm$ 0.003 \\
             FS-TKY & 0.989 $\pm$ 0.002 \\
             Rotten\_tomatoes & 0.960 $\pm$ 0.011 \\
        \bottomrule
        \end{tabular} }
        \caption{NDCG on predicted Pareto fronts across all runs and weight configurations \([\omega_A, \omega_E]\), where \(\omega_A + \omega_E = 1\).}
        \label{tab:ndcg_summary}
    \end{minipage}
    \hfill
    \begin{minipage}{0.55\textwidth}
        \centering
         \resizebox{\textwidth}{!}{
        \begin{tabular}{cccc}
        \toprule
             Dataset & HaD ($\downarrow$) & $\Delta HV$  ($\downarrow$)\\
        \midrule
             CIFAR-10 & 0.050 $\pm$ 0.011 & 0.009 $\pm$ 0.006 \\
             FS-TKY & 0.075 $\pm$ 0.002 & 0.049 $\pm$ 0.030 \\
             Rotten\_tomatoes & 0.312 $\pm$ 0.021 & 0.195 $\pm$ 0.060 \\
        \bottomrule
        \end{tabular}}
        \caption{Hausdorff Distance (HaD) and Hypervolume Difference ($\Delta HV$) on the 3 test sets.}
        \label{tab:Haus_HV_2}
    \end{minipage}
\end{table}

\section{Results}
\label{sec:results}

\begin{figure*}
    \centering
\includegraphics[width=\linewidth]{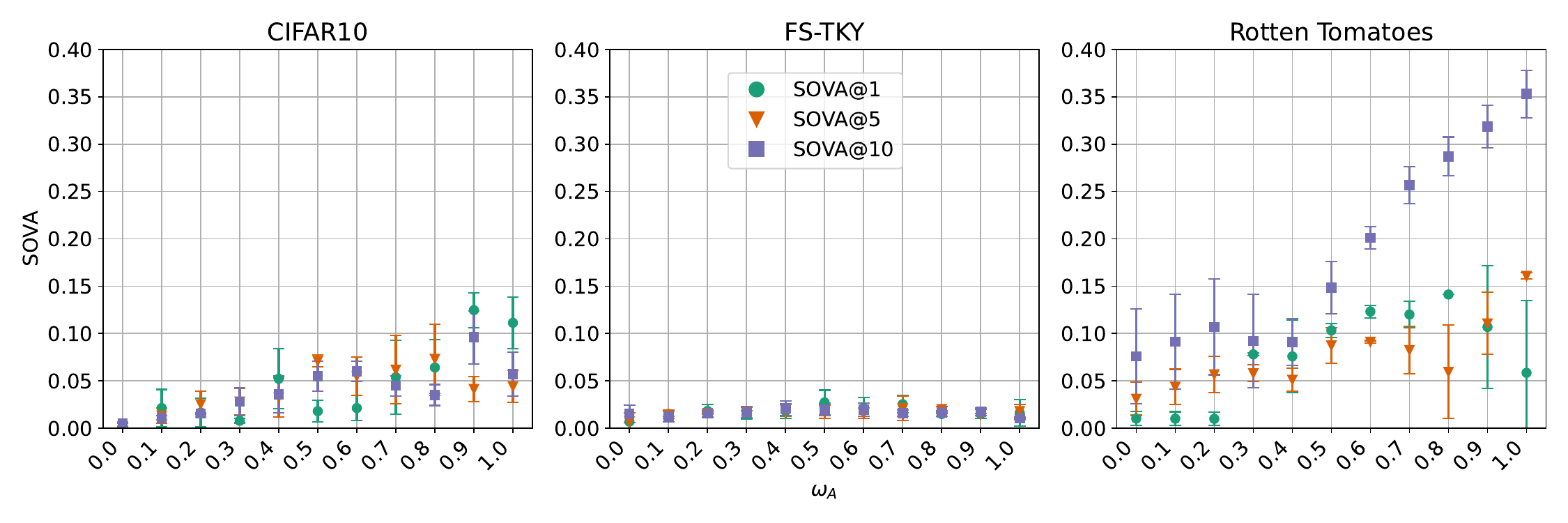}
    \caption{Mean and standard deviation of SOVA@k across test datasets at varying  \( \omega_A \). \( \omega_A \) represents the weight assigned to the validation accuracy target relative to the energy target \((\omega_A +\omega_E)=1\).}
    \label{fig:sova}
\end{figure*}

\begin{table*}[ht]
    \centering
    \resizebox{\linewidth}{!}{
    \begin{tabular}{cccccccc}
    \toprule
         Dataset & Recall$_{\text{EE}}$ ($\uparrow$) & Recall$_{\text{RE}}$ ($\uparrow$) & Recall$_{\text{IE}}$ ($\uparrow$) & F1$_{\text{EE}}$ ($\uparrow$) & F1$_{\text{RE}}$ ($\uparrow$) & F1$_{\text{IE}}$ ($\uparrow$)  \\
         \midrule
         CIFAR-10 & 0.367 $\pm$ 0.462 & 0.574 $\pm$ 0.302 & 0.971 $\pm$ 0.025 & 0.008 $\pm$ 0.011 & 0.059 $\pm$ 0.081 & 0.186 $\pm$ 0.106 \\
         FS-TKY & 0.000 $\pm$ 0.000 & 0.023 $\pm$ 0.040 & 0.995 $\pm$ 0.008 & 0.000 $\pm$ 0.000 & 0.002 $\pm$ 0.003 & 0.388 $\pm$ 0.236  \\
         Rotten\_tomatoes & 0.467 $\pm$ 0.115 & 0.894 $\pm$ 0.094 & 0.917 $\pm$ 0.144 & 0.142 $\pm$ 0.028 & 0.692 $\pm$ 0.068 & 0.532 $\pm$ 0.075 \\
         \bottomrule
    \end{tabular}
    }
    \caption{Mean $\pm$ Std of  performance metrics for evaluating \approach on test datasets. Recall and F1 are reported under three scenarios: Exact Epoch (EE), Relaxed Epoch (RE, ±5 epochs), and Ignored Epoch (IE, no epoch constraints).}
    \label{tab:ranking_metrics}
\end{table*}

\textbf{Prediction Accuracy.}
 \cref{tab:mae_performance} highlights the performance of our predictive model, showing the mean and standard deviation of the MAE achieved across three independent runs on the test datasets, with different percentages of samples discarded. We observe that the accuracy of predictions is more sensitive to increasing discard rates compared to energy predictions, particularly for CIFAR-10 and Rotten\_tomatoes. This suggests that as the complexity of the task increases due to modifications in the dataset, the learning behavior of the model configurations across training epochs becomes less predictable. The relatively lower predictive performance observed on the Rotten\_tomatoes dataset is likely due to the specific nature of the model architectures considered in the NLP domain, particularly their dimensionality, along with the lower number of tracked training dynamics and the shorter sequence of training epochs for NLP experiments used to train the predictor model, compared to experiments in other domains tracked in \dname (see Table \ref{tab:knowledge_base_summary}).Further sanity checks assessing the robustness of our predictive pipeline are provided in \cref{app:additional_studies}, while disaggregated results can be found in \cref{app:additional_results}.

\textbf{Robustness and Effectiveness of Ranking.}
Given that our ultimate goal is to recommend Pareto-optimal combinations of model architecture, batch size, learning rate, and number of training epochs that closely align with the true Pareto front, ensuring consistent and reliable rankings of model configurations based on user preferences is more critical than achieving perfect accuracy in predicting the target metrics. Even with minor prediction errors, in fact, as long as these errors are systematically consistent with the true values, their overall impact remains minimal. This is evidenced by the high average NDCG scores reported in \cref{tab:ndcg_summary}, which demonstrates strong alignment between the predicted and actual rankings across different weight configurations that modulate the relative importance of the two objectives in determining the ranking score. The weight configurations include different combinations of \(\omega_A\) and \(\omega_E\), with values ranging between 0 and 1 in steps of 0.1. These results highlight the robustness of \approach in selecting and ranking different model configurations for a given problem setting, even when preferences vary regarding the prioritization of objectives. 

\textbf{Pareto Front Alignment.}
\cref{tab:ranking_metrics}, instead, provides quantitative insights into the quality of the predicted Pareto front by showcasing the Recall and F1-scores under different evaluation settings. Specifically, the table presents metrics computed under three evaluation scenarios: the \textit{Exact Epoch (EE)} scenario assesses the alignment between the true and predicted Pareto fronts by considering both the model configuration and the exact number of training epochs; the \textit{Relaxed Epoch (RE)} scenario allows for minor deviations in epoch selection (\(\pm 5\) epochs) while still evaluating the compatibility of the Pareto fronts; and the \textit{Ignored Epoch (IE)} scenario evaluates the overlap of items in the Pareto fronts without considering the suggested number of training epochs.
The results indicate that under the EE setting—where both the model configuration and the exact training epoch must be correctly predicted—the Recall and F1 scores are generally lower, particularly for the FS-TKY dataset, which shows minimal overlap due to its wider epoch space (400 epochs) compared to the CV  and NLP domains, where the epoch sequences has respectively length 100 and 5. It should be emphasized, however, that this level of granularity in recommendations— 
particularly concerning the precise alignment of model configurations and training epochs— is rarely addressed in most NAS or model selection approaches. 
Significant improvements in Pareto matching performance are observed when transitioning to the RE setting and, even more so, to the  IE setting, demonstrating the effectiveness of \approach in accurately identifying configurations that are truly Pareto-optimal. 
\cref{tab:Haus_HV_2} reports the Hausdorff Distance (HaD) and Hypervolume Difference (\( \Delta HV \)) across the three test datasets, offering complementary views on spatial deviation and coverage between the predicted and true Pareto fronts. In general, both metrics yield values close to zero, indicating strong alignment and effective approximation of the true fronts. The slightly higher HaD and \( \Delta HV \) values observed for the \textit{Rotten\_tomatoes} dataset is consistent with the less accurate underlying predictions for the objective metrics in that setting. Overall, these results highlight the effectiveness of \approach in accurately capturing the structure and composition of the Pareto front across various datasets and problem settings, while also highlighting the sensitivity of front reconstruction to the effectiveness of the target prediction task. Due to space constraints, a visual comparison of the predicted and true Pareto frontiers is presented in \cref{app:additional_studies} as complementary material.

\textbf{Consistency of Ranked Pareto-optimal Solutions.}
The SOVA@k results, presented in \cref{fig:sova}, provide an in-depth analysis of \approach's ability to maintain ranking consistency between the predicted and true Pareto fronts across different datasets and varying objective weights (represented as \( \omega_A \) on the x-axis).
 For CIFAR-10, the SOVA@1, SOVA@5, and SOVA@10 scores remain relatively low and stable across most values of \( \omega_A \), demonstrating consistent alignment of rankings. However, a slight increase in SOVA@k values is observed as \( \omega_A \) approaches 1, indicating minor performance degradation when the priority shifts solely toward maximizing performance, regardless of energy consumption. The FS-TKY dataset displays consistently low and stable SOVA@k scores across all settings, suggesting that \approach effectively preserves ranking consistency in this domain, even under diverse weight configurations. In contrast, the Rotten\_tomatoes dataset reveals an upward trend in SOVA@k scores—particularly SOVA@10—as \( \omega_A \) increases. This degradation is closely tied to the comparatively higher MAE in validation accuracy for Rotten\_tomatoes (reported in Table~\ref{tab:mae_performance}), where even small prediction errors can lead to substantial ranking misalignments\footnote{To better understand this, recall that SOVA@k computes a weighted sum of absolute differences in the true objective values of matched configurations.} 
 
\textbf{Comparison with Competitors.} 
To the best of our knowledge, no existing method directly addresses the goal of energy-aware, cross-domain model selection over standard architectures. To contextualize our results, however, we compare our approach with two representative Eco-NAS baselines: EC-NAS and KNAS. Notably, these methods are specifically designed for Eco-NAS within constrained architecture spaces composed of closely related models, whereas \approach targets a broader scenario, aiming to generalize across datasets and architecture families.
Despite considerable effort, it was not possible to adapt the codebases of these baselines to our cross-domain search space. For this reason, we provide both an illustrative and a more NAS-specific quantitative comparison by presenting their results on the NAS benchmarks originally used in their respective publications—NASBench-101 for EC-NAS and NASBench-201 for KNAS.
As part of the illustrative comparison, \cref{tab:competitors} reports the predicted performance of the Pareto-optimal configurations suggested by each method—both when the objective is to maximize performance (\_MA) and when equal importance is given to performance and energy consumption (\_B), along with the runtime required to produce each solution. As shown in the table, \approach consistently recommends configurations that achieve strong trade-offs between validation accuracy and energy usage. Moreover, a notable advantage of \approach lies in its efficiency: although training the predictive model incurs a one-time computational cost, inference is extremely fast. In contrast, the Eco-NAS baselines require re-running the full optimization or search process for every new dataset or constraint, resulting in significantly higher computational overhead. In the second comparative setting, we evaluate the behavior of \approach 
within a NAS-specific benchmark. Specifically, we assess its performance on NASBench-101, enabling a fair comparison with EC-NAS and showcasing its capacity to generalize beyond its original cross-domain design. As shown in \cref{tab:competitors_NAS}, although \approach  was not originally designed for NAS tasks, it nonetheless demonstrates the ability to operate effectively within such constrained settings.  Enhancing the performance of \approach in NAS-specific contexts—through the development of refined feature representations that better capture the subtle architectural distinctions characteristic of NASBench-style benchmarks—is left to future work.


\begin{table}[ht]
    \centering
    \begin{minipage}
    {0.5\textwidth}
    \centering
    \resizebox{\textwidth}{!}{
    \begin{tabular}{cccc}
    \toprule
    Method & Predicted A (acc) & Predicted E (kWh) & Time (s) \\
    \midrule
    EC-NAS$_{\text{MA}}$ & 0.822 (-0.148) & 27.745 (27.673) & 564\\
    EC-NAS$_{\text{B}}$ & 0.771 (-0.191) & 8.827 (8.814) & 564\\
    KNAS &  0.183 (-0.787) & 0.526 (0.454) & 27,960\\
    \approach$_{\text{MA}}$ (ours) & \textbf{0.899} 
 (-0.071) & 0.509 
 (0.437) & 1,241+\textbf{12}\\
 \approach$_{\text{B}}$ (ours) & 0.887 
 \textbf{(-0.075)} & \textbf{0.086} 
 \textbf{(0.073)} & 1,241+\textbf{12}\\
 \bottomrule
    \end{tabular} }
    \caption{Comparison of \approach vs. competitors in accuracy (A), energy (E), and computational time (s). Brackets show the gap between suggested configs and the best ground truth in \dname. In \textbf{bold} is highlighted the best result for each column. Bold value after $+$ for \approach shows inference time, as training occurs once.}
    \label{tab:competitors}
    \end{minipage}
    \hfill
    \begin{minipage}{0.45\textwidth}
    \centering
    \resizebox{\textwidth}{!}{
    \begin{tabular}{lcc}
    \toprule
    \textbf{Method} & \textbf{Validation Accuracy} & \textbf{Training Time (s)} \\
    \midrule
    EC-NAS & \textbf{0.946} $(-0.5\%)$ & 3160 $(-34.1\%)$ \\
    \approach (ours) & 0.917 $(-3.5\%)$ & \textbf{1628} $(-66.0\%)$\\
    \bottomrule
    \end{tabular} }
    \caption{Comparison of \approach and EC-NAS in terms of ground-truth validation accuracy and training time (in seconds), as reported in NASBench-101. Each solution corresponds to the predicted Pareto-optimal configuration maximizing validation accuracy at epoch 108.   The values, shown in \textbf{bold} in the table represent the best solution for each individual objective. 
    }
    \label{tab:competitors_NAS}
    \end{minipage}
\end{table}

\section{Conclusions and Future Work}
\label{sec:conclusion}

This work addresses the critical challenge of environmental sustainability in AI development by introducing a novel approach to eco-efficient model selection and optimization. Our method offers a flexible, domain-agnostic solution for recommending Pareto-optimal NN configurations that balance performance and energy consumption.
Operating at inference time, our approach overcomes the limitations of traditional NAS and HPO, demonstrating effectiveness across diverse AI domains. 

The release of \dname provides researchers and practitioners with valuable resources to advance eco-efficient machine learning. We hope that our work contributes to a more sustainable future by enabling informed decisions that consider performance and energy efficiency.

\looseness -1 Future work aims to develop a framework that automatically updates the knowledge base with new  experiments, enabling \dname to expand to various tasks without manual intervention.  \

\bibliography{main}

\begin{thebibliography}{}

\bibitem[Bakhtiarifard et~al., 2024]{ec_nas}
Bakhtiarifard, P., Igel, C., and Selvan, R. (2024).
\newblock Ec-nas: Energy consumption aware tabular benchmarks for neural architecture search.
\newblock In {\em ICASSP 2024-2024 IEEE International Conference on Acoustics, Speech and Signal Processing (ICASSP)}, pages 5660--5664. IEEE.

\bibitem[Bender et~al., 2021]{10.1145/3442188.3445922}
Bender, E.~M., Gebru, T., McMillan-Major, A., and Shmitchell, S. (2021).
\newblock On the dangers of stochastic parrots: Can language models be too big?
\newblock In {\em Proceedings of the 2021 ACM Conference on Fairness, Accountability, and Transparency}, FAccT '21, page 610–623, New York, NY, USA. Association for Computing Machinery.

\bibitem[Betello et~al., 2024]{betello2024reproducible}
Betello, F., Purificato, A., Siciliano, F., Trappolini, G., Bacciu, A., Tonellotto, N., and Silvestri, F. (2024).
\newblock A reproducible analysis of sequential recommender systems.
\newblock {\em IEEE Access}.

\bibitem[Bossard et~al., 2014]{bossard2014food}
Bossard, L., Guillaumin, M., and Van~Gool, L. (2014).
\newblock Food-101--mining discriminative components with random forests.
\newblock In {\em Computer vision--ECCV 2014: 13th European conference, zurich, Switzerland, September 6-12, 2014, proceedings, part VI 13}, pages 446--461. Springer.

\bibitem[Chung et~al., 2024]{chung2024reducing}
Chung, J.-W., Gu, Y., Jang, I., Meng, L., Bansal, N., and Chowdhury, M. (2024).
\newblock Reducing energy bloat in large model training.
\newblock In {\em Proceedings of the ACM SIGOPS 30th Symposium on Operating Systems Principles}, pages 144--159.

\bibitem[Courty et~al., 2023]{codecarbon}
Courty, B., Schmidt, V., Goyal-Kamal, Coutarel, M., Feld, B., Lecourt, J., SabAsmine, kngoyal, Léval, M., Cruveiller, A., inimaz, ouminasara, Zhao, F., Joshi, A., Bogroff, A., Saboni, A., de~Lavoreille, H., Laskaris, N., Blanche, L., Abati, E., LiamConnell, Blank, D., Wang, Z., Catovic, A., St{k{e}}ch{l}y, M., alencon, JPW, MinervaBooks, \c{C}arkac\i{}, N., and DomAlexRod (2023).
\newblock mlco2/codecarbon: v2.3.2.

\bibitem[Dale and Chall, 1948]{dale1948formula}
Dale, E. and Chall, J.~S. (1948).
\newblock A formula for predicting readability: Instructions.
\newblock {\em Educational research bulletin}, pages 37--54.

\bibitem[DeepSeek-AI, 2024]{deepseek}
DeepSeek-AI (2024).
\newblock Deepseek-v3 technical report.

\bibitem[Devlin et~al., 2019]{devlinetal2019bert}
Devlin, J., Chang, M.-W., Lee, K., and Toutanova, K. (2019).
\newblock {BERT}: Pre-training of deep bidirectional transformers for language understanding.
\newblock In Burstein, J., Doran, C., and Solorio, T., editors, {\em Proceedings of the 2019 Conference of the North {A}merican Chapter of the Association for Computational Linguistics: Human Language Technologies, Volume 1 (Long and Short Papers)}, pages 4171--4186, Minneapolis, Minnesota. Association for Computational Linguistics.

\bibitem[Dong and Yang, 2020]{dong2020bench}
Dong, X. and Yang, Y. (2020).
\newblock Nas-bench-201: Extending the scope of reproducible neural architecture search.
\newblock {\em arXiv preprint arXiv:2001.00326}.

\bibitem[Dosovitskiy, 2020]{dosovitskiy2020image}
Dosovitskiy, A. (2020).
\newblock An image is worth 16x16 words: Transformers for image recognition at scale.
\newblock {\em arXiv preprint arXiv:2010.11929}.

\bibitem[Dou et~al., 2023]{ea_has_bench}
Dou, S., Jiang, X., Zhao, C.~R., and Li, D. (2023).
\newblock Ea-has-bench: Energy-aware hyperparameter and architecture search benchmark.
\newblock In {\em The Eleventh International Conference on Learning Representations}.

\bibitem[Elsken et~al., 2018]{elsken2018efficient}
Elsken, T., Metzen, J.~H., and Hutter, F. (2018).
\newblock Efficient multi-objective neural architecture search via lamarckian evolution.
\newblock {\em arXiv preprint arXiv:1804.09081}.

\bibitem[Faiz et~al., 2024]{faiz2024llmcarbon}
Faiz, A., Kaneda, S., Wang, R., Osi, R., Sharma, P., Chen, F., and Jiang, L. (2024).
\newblock Llmcarbon: Modeling the end-to-end carbon footprint of large language models.
\newblock In {\em The Twelfth International Conference on Learning Representations}. ICLR.

\bibitem[George et~al., 2023]{george2023environmental}
George, A.~S., George, A.~H., and Martin, A.~G. (2023).
\newblock The environmental impact of ai: a case study of water consumption by chat gpt.
\newblock {\em Partners Universal International Innovation Journal}, 1(2):97--104.

\bibitem[Guo et~al., 2020]{guo2020breaking}
Guo, Y., Chen, Y., Zheng, Y., Zhao, P., Chen, J., Huang, J., and Tan, M. (2020).
\newblock Breaking the curse of space explosion: Towards efficient nas with curriculum search.
\newblock In {\em International Conference on Machine Learning}, pages 3822--3831. PMLR.

\bibitem[Harper and Konstan, 2015]{10.1145/2827872}
Harper, F.~M. and Konstan, J.~A. (2015).
\newblock The movielens datasets: History and context.
\newblock {\em ACM Trans. Interact. Intell. Syst.}, 5(4).

\bibitem[He et~al., 2016]{he2016deep}
He, K., Zhang, X., Ren, S., and Sun, J. (2016).
\newblock Deep residual learning for image recognition.
\newblock In {\em Proceedings of the IEEE conference on computer vision and pattern recognition}, pages 770--778.

\bibitem[Henrikson, 1999]{henrikson1999completeness}
Henrikson, J. (1999).
\newblock Completeness and total boundedness of the hausdorff metric.
\newblock {\em MIT Undergraduate Journal of Mathematics}, 1(69-80):10.

\bibitem[Hidasi et~al., 2016]{hidasi2016sessionbased}
Hidasi, B., Karatzoglou, A., Baltrunas, L., and Tikk, D. (2016).
\newblock Session-based recommendations with recurrent neural networks.

\bibitem[Hou et~al., 2022]{hou2022core}
Hou, Y., Hu, B., Zhang, Z., and Zhao, W.~X. (2022).
\newblock Core: simple and effective session-based recommendation within consistent representation space.

\bibitem[Iandola, 2016]{iandola2016squeezenet}
Iandola, F.~N. (2016).
\newblock Squeezenet: Alexnet-level accuracy with 50x fewer parameters and< 0.5 mb model size.
\newblock {\em arXiv preprint arXiv:1602.07360}.

\bibitem[Javaheripi et~al., 2023]{javaheripi2023phi}
Javaheripi, M., Bubeck, S., Abdin, M., Aneja, J., Bubeck, S., Mendes, C. C.~T., Chen, W., Del~Giorno, A., Eldan, R., Gopi, S., et~al. (2023).
\newblock Phi-2: The surprising power of small language models.
\newblock {\em Microsoft Research Blog}.

\bibitem[Jiang et~al., 2023]{jiang2023mistral}
Jiang, A.~Q., Sablayrolles, A., Mensch, A., Bamford, C., Chaplot, D.~S., Casas, D. d.~l., Bressand, F., Lengyel, G., Lample, G., Saulnier, L., et~al. (2023).
\newblock Mistral 7b.
\newblock {\em arXiv preprint arXiv:2310.06825}.

\bibitem[Kang and McAuley, 2018]{kang2018selfattentive}
Kang, W.-C. and McAuley, J. (2018).
\newblock Self-attentive sequential recommendation.

\bibitem[Kincaid, 1975]{kincaid1975derivation}
Kincaid, J. (1975).
\newblock Derivation of new readability formulas (automated readability index, fog count and flesch reading ease formula) for navy enlisted personnel.
\newblock {\em Chief of Naval Technical Training}.

\bibitem[Krizhevsky et~al., 2009]{krizhevsky2009learning}
Krizhevsky, A., Hinton, G., et~al. (2009).
\newblock Learning multiple layers of features from tiny images.

\bibitem[Krizhevsky et~al., 2012]{krizhevsky2012imagenet}
Krizhevsky, A., Sutskever, I., and Hinton, G.~E. (2012).
\newblock Imagenet classification with deep convolutional neural networks.
\newblock {\em Advances in neural information processing systems}, 25.

\bibitem[LeCun et~al., 2010]{lecun2010mnist}
LeCun, Y., Cortes, C., and Burges, C. (2010).
\newblock Mnist handwritten digit database.
\newblock {\em ATT Labs [Online]. Available: http://yann.lecun.com/exdb/mnist}, 2.

\bibitem[Li and Yao, 2019]{li2019quality}
Li, M. and Yao, X. (2019).
\newblock Quality evaluation of solution sets in multiobjective optimisation: A survey.
\newblock {\em ACM Computing Surveys (CSUR)}, 52(2):1--38.

\bibitem[Liu et~al., 2018]{liu2018darts}
Liu, H., Simonyan, K., and Yang, Y. (2018).
\newblock Darts: Differentiable architecture search.
\newblock {\em arXiv preprint arXiv:1806.09055}.

\bibitem[Liu et~al., 2022]{liu2022survey}
Liu, S., Zhang, H., and Jin, Y. (2022).
\newblock A survey on computationally efficient neural architecture search.
\newblock {\em Journal of Automation and Intelligence}, 1(1):100002.

\bibitem[Liu, 2019]{liu2019roberta}
Liu, Y. (2019).
\newblock Roberta: A robustly optimized bert pretraining approach.
\newblock {\em arXiv preprint arXiv:1907.11692}, 364.

\bibitem[Metz et~al., 2020]{metz2020using}
Metz, L., Maheswaranathan, N., Sun, R., Freeman, C.~D., Poole, B., and Sohl-Dickstein, J. (2020).
\newblock Using a thousand optimization tasks to learn hyperparameter search strategies.
\newblock {\em arXiv preprint arXiv:2002.11887}.

\bibitem[Pang and Lee, 2005]{PangLee05a}
Pang, B. and Lee, L. (2005).
\newblock Seeing stars: Exploiting class relationships for sentiment categorization with respect to rating scales.
\newblock In {\em Proceedings of the ACL}.

\bibitem[Pareto, 1964]{pareto1964cours}
Pareto, V. (1964).
\newblock {\em Cours d'{\'e}conomie politique}, volume~1.
\newblock Librairie Droz.

\bibitem[Saravia et~al., 2018]{saraviaetal2018carer}
Saravia, E., Liu, H.-C.~T., Huang, Y.-H., Wu, J., and Chen, Y.-S. (2018).
\newblock {CARER}: Contextualized affect representations for emotion recognition.
\newblock In {\em Proceedings of the 2018 Conference on Empirical Methods in Natural Language Processing}, pages 3687--3697, Brussels, Belgium. Association for Computational Linguistics.

\bibitem[Schutze et~al., 2012]{6151115}
Schutze, O., Esquivel, X., Lara, A., and Coello, C. A.~C. (2012).
\newblock Using the averaged hausdorff distance as a performance measure in evolutionary multiobjective optimization.
\newblock {\em IEEE Transactions on Evolutionary Computation}, 16(4):504--522.

\bibitem[Siems et~al., 2020]{siems2020bench}
Siems, J., Zimmer, L., Zela, A., Lukasik, J., Keuper, M., and Hutter, F. (2020).
\newblock Nas-bench-301 and the case for surrogate benchmarks for neural architecture search.
\newblock {\em arXiv preprint arXiv:2008.09777}, 4:14.

\bibitem[Simonyan and Zisserman, 2014]{simonyan2014very}
Simonyan, K. and Zisserman, A. (2014).
\newblock Very deep convolutional networks for large-scale image recognition.
\newblock {\em arXiv preprint arXiv:1409.1556}.

\bibitem[Strubell et~al., 2020]{strubell2020energy}
Strubell, E., Ganesh, A., and McCallum, A. (2020).
\newblock Energy and policy considerations for modern deep learning research.
\newblock In {\em Proceedings of the AAAI conference on artificial intelligence}, volume~34, pages 13693--13696.

\bibitem[Sun et~al., 2019]{sun2019bert4rec}
Sun, F., Liu, J., Wu, J., Pei, C., Lin, X., Ou, W., and Jiang, P. (2019).
\newblock Bert4rec: Sequential recommendation with bidirectional encoder representations from transformer.

\bibitem[Tan and Le, 2019]{tan2019efficientnet}
Tan, M. and Le, Q. (2019).
\newblock Efficientnet: Rethinking model scaling for convolutional neural networks.
\newblock In {\em International conference on machine learning}, pages 6105--6114. PMLR.

\bibitem[Touvron et~al., 2023]{touvron2023llama2openfoundation}
Touvron, H., Martin, L., Stone, K., Albert, P., Almahairi, A., Babaei, Y., Bashlykov, N., Batra, S., Bhargava, P., Bhosale, S., Bikel, D., Blecher, L., Ferrer, C.~C., Chen, M., Cucurull, G., Esiobu, D., Fernandes, J., Fu, J., Fu, W., Fuller, B., Gao, C., Goswami, V., Goyal, N., Hartshorn, A., Hosseini, S., Hou, R., Inan, H., Kardas, M., Kerkez, V., Khabsa, M., Kloumann, I., Korenev, A., Koura, P.~S., Lachaux, M.-A., Lavril, T., Lee, J., Liskovich, D., Lu, Y., Mao, Y., Martinet, X., Mihaylov, T., Mishra, P., Molybog, I., Nie, Y., Poulton, A., Reizenstein, J., Rungta, R., Saladi, K., Schelten, A., Silva, R., Smith, E.~M., Subramanian, R., Tan, X.~E., Tang, B., Taylor, R., Williams, A., Kuan, J.~X., Xu, P., Yan, Z., Zarov, I., Zhang, Y., Fan, A., Kambadur, M., Narang, S., Rodriguez, A., Stojnic, R., Edunov, S., and Scialom, T. (2023).
\newblock Llama 2: Open foundation and fine-tuned chat models.

\bibitem[Vente et~al., 2024]{from_clicks_to_carbon}
Vente, T., Wegmeth, L., Said, A., and Beel, J. (2024).
\newblock From clicks to carbon: The environmental toll of recommender systems.
\newblock In {\em Proceedings of the 18th ACM Conference on Recommender Systems}, RecSys '24, page 580–590, New York, NY, USA. Association for Computing Machinery.

\bibitem[Wang et~al., 2013]{wang2013theoretical}
Wang, Y., Wang, L., Li, Y., He, D., and Liu, T.-Y. (2013).
\newblock A theoretical analysis of ndcg type ranking measures.
\newblock In {\em Conference on learning theory}, pages 25--54. PMLR.

\bibitem[Wang et~al., 2020]{wang2020benchmarking}
Wang, Y., Wang, Q., Shi, S., He, X., Tang, Z., Zhao, K., and Chu, X. (2020).
\newblock Benchmarking the performance and energy efficiency of ai accelerators for ai training.
\newblock In {\em 2020 20th IEEE/ACM International Symposium on Cluster, Cloud and Internet Computing (CCGRID)}, pages 744--751. IEEE.

\bibitem[Wu et~al., 2022]{wu2022sustainable}
Wu, C.-J., Raghavendra, R., Gupta, U., Acun, B., Ardalani, N., Maeng, K., Chang, G., Aga, F., Huang, J., Bai, C., et~al. (2022).
\newblock Sustainable ai: Environmental implications, challenges and opportunities.
\newblock {\em Proceedings of Machine Learning and Systems}, 4:795--813.

\bibitem[Xiao et~al., 2017]{xiao2017fashion}
Xiao, H., Rasul, K., and Vollgraf, R. (2017).
\newblock Fashion-mnist: a novel image dataset for benchmarking machine learning algorithms.
\newblock {\em arXiv preprint arXiv:1708.07747}.

\bibitem[Xu et~al., 2021]{xu2021knas}
Xu, J., Zhao, L., Lin, J., Gao, R., Sun, X., and Yang, H. (2021).
\newblock Knas: green neural architecture search.
\newblock In {\em International Conference on Machine Learning}, pages 11613--11625. PMLR.

\bibitem[Yang et~al., 2014]{yang2014modeling}
Yang, D., Zhang, D., Zheng, V.~W., and Yu, Z. (2014).
\newblock Modeling user activity preference by leveraging user spatial temporal characteristics in lbsns.
\newblock {\em IEEE Transactions on Systems, Man, and Cybernetics: Systems}, 45(1):129--142.

\bibitem[Ying et~al., 2019]{ying2019bench}
Ying, C., Klein, A., Christiansen, E., Real, E., Murphy, K., and Hutter, F. (2019).
\newblock Nas-bench-101: Towards reproducible neural architecture search.
\newblock In {\em International conference on machine learning}, pages 7105--7114. PMLR.

\bibitem[You et~al., 2023]{you2023zeus}
You, J., Chung, J.-W., and Chowdhury, M. (2023).
\newblock Zeus: Understanding and optimizing $\{$GPU$\}$ energy consumption of $\{$DNN$\}$ training.
\newblock In {\em 20th USENIX Symposium on Networked Systems Design and Implementation (NSDI 23)}, pages 119--139.

\bibitem[Zela et~al., 2020]{zela2020surrogate}
Zela, A., Siems, J., Zimmer, L., Lukasik, J., Keuper, M., and Hutter, F. (2020).
\newblock Surrogate nas benchmarks: Going beyond the limited search spaces of tabular nas benchmarks.
\newblock {\em arXiv preprint arXiv:2008.09777}.

\bibitem[Zhao et~al., 2024]{ce_nas}
Zhao, Y., Liu, Y., Jiang, B., and Guo, T. (2024).
\newblock {CE}-{NAS}: An end-to-end carbon-efficient neural architecture search framework.
\newblock In {\em The Thirty-eighth Annual Conference on Neural Information Processing Systems}.

\bibitem[Zimmer et~al., 2021]{ZimLin2021a}
Zimmer, L., Lindauer, M., and Hutter, F. (2021).
\newblock Auto-pytorch tabular: Multi-fidelity metalearning for efficient and robust autodl.
\newblock {\em IEEE Transactions on Pattern Analysis and Machine Intelligence}, 43(9):3079 -- 3090.

\bibitem[Zitzler and Thiele, 1998]{zitzler1998multiobjective}
Zitzler, E. and Thiele, L. (1998).
\newblock Multiobjective optimization using evolutionary algorithms—a comparative case study.
\newblock In {\em International conference on parallel problem solving from nature}, pages 292--301. Springer.

\end{thebibliography}
\bibliographystyle{apalike}


\appendix

\section{Technical Appendices and Supplementary Material}
\subsection{Overview of Datasets and Models}
\label{appendix:d_m_details}
\paragraph{Vision Models} 

\begin{itemize}
    \item \textbf{AlexNet} \citep{krizhevsky2012imagenet}: One of the first CNNs, known for its 8-layer architecture, performed well in large-scale image classification.
    \item \textbf{EfficientNet} \citep{tan2019efficientnet}: A family of CNN that balances accuracy and efficiency by systematically scaling width, depth and resolution.
    \item \textbf{ResNet18} \citep{he2016deep}: 18-layer lightweight residual NN using skip connections to solve the vanishing gradient problem.
    \item \textbf{SqueezeNet} \citep{iandola2016squeezenet}: An ultra-lightweight convolutional NN designed for model size efficiency with fire modules for parameter reduction.
    \item  \textbf{ViT} \citep{dosovitskiy2020image}: A transformer-based architecture that applies self-attention mechanisms to image patches for superior image recognition.
    \item \textbf{VGG16} \citep{simonyan2014very}: A deep convolutional NN with 16 layers, known for its simplicity and uniform use of 3x3 convolutional filters.
\end{itemize}

\paragraph{Vision Datasets}
\begin{itemize}
    \item \textbf{CIFAR-10} \citep{krizhevsky2009learning}: It consists of $60,000$ $32x32$ color images divided in 10 classes, with $6,000$ images per class. There are $50,000$ training images and $10,000$ test images.
    \item \textbf{FOOD101} \citep{bossard2014food}: It comprises 101 food categories with $750$ training and $250$ test images per category, for a total of $101K$ images.
    \item \textbf{MNIST} \citep{lecun2010mnist}: It is a large collection of handwritten digits. It has a training set of $60,000$ examples, and a test set of $10,000$ examples.
    \item \textbf{Fashion-MNIST} \citep{xiao2017fashion}: It consists of $28x28$ greyscale images of $70,000$ fashion products from 10 categories, with $7,000$ images per category. The training set has $60,000$ images and the test set has $10,000$ images.
\end{itemize} 

\paragraph{Text Models}

\begin{itemize}
    \item \textbf{RoBERTa} \citep{liu2019roberta}: An optimized version of BERT by Facebook that improves performance through larger datasets and longer training.
    \item \textbf{BERT} \citep{devlinetal2019bert}: A groundbreaking transformer-based model by Google that uses bidirectional attention to understand the context of words in a sentence.
    \item \textbf{Microsoft-PHI-2} \citep{javaheripi2023phi}: A small LLM specialized model by Microsoft, a Transformer with 2.7 billion parameters.
    \item \textbf{Mistral-7B-v0.3} \citep{jiang2023mistral}: A highly efficient, open-weight, 7-billion-parameter language model offering strong performance in text generation and understanding tasks.
\end{itemize}


\paragraph{Text Datasets}

\begin{itemize}
    \item \textbf{Google-boolq}\footnote{\url{https://huggingface.co/datasets/google/boolq}}: It's a question answering dataset for yes/no questions containing 15942 example. Each example is a triplet of (question, passage, answer).
    \item \textbf{StanfordNLP-IMDB}\footnote{\url{https://huggingface.co/datasets/stanfordnlp/imdb}}: This is a dataset for binary sentiment classification. They provide a set of 25,000 highly polar movie reviews for training, and 25,000 for testing.
    \item \textbf{Dair-ai/Emotions}\footnote{\url{https://huggingface.co/datasets/dair-ai/emotion}} \citep{saraviaetal2018carer}: It is a dataset of English Twitter messages with six basic emotions: anger, fear, joy, love, sadness, and surprise.
    \item \textbf{Rotten\_tomatoes}\footnote{\url{https://huggingface.co/datasets/cornell-movie-review-data/rotten_tomatoes}} \citep{PangLee05a}: This is a dataset of containing 5,331 positive and 5,331 negative processed sentences from Rotten\_tomatoes movie reviews.
\end{itemize} 

\paragraph{Recommendation Models}

\begin{itemize}
    \item  \textbf{BERT4Rec} \citep{sun2019bert4rec}: This model is based on the BERT architecture, enabling it to capture complex relationships in user behaviour sequences through bidirectional self-attention.
    \item \textbf{CORE} \citep{hou2022core}: it introduces an attention mechanism that enables the model to weigh the contribution of each item in the input sequence, enhancing recommendation accuracy.
    \item \textbf{GRU4Rec} \citep{hidasi2016sessionbased}: This model utilizes GRUs to capture temporal dependencies in user-item interactions.
    \item \textbf{SASRec} \citep{kang2018selfattentive}: This model is characterized by its use of self-attention mechanisms, allowing it to discern the relevance of each item within the user's sequence.
\end{itemize}

\paragraph{Recommendation Datasets}
\begin{itemize}
    \item \textbf{Foursquare}\footnote{\url{https://sites.google.com/site/yangdingqi/home/foursquare-dataset}}: These datasets contain check-ins collected over a period of approximately ten months \citep{yang2014modeling}. We use the New York City (FS-NYC) and Tokyo (FS-TKY) versions.
    \item \textbf{MovieLens}\footnote{\url{https://grouplens.org/datasets/movielens}}: The MovieLens dataset \citep{10.1145/2827872} is widely recognized as a benchmark for evaluating recommendation algorithms. We utilize two versions: MovieLens 1M (ML-1M) and MovieLens 100k (ML-100k).
\end{itemize}
Our pre-processing approach adheres to common practices, where ratings are treated as implicit feedback, meaning all interactions are utilized regardless of their rating values, and users or items with fewer than five interactions are excluded \citep{kang2018selfattentive,sun2019bert4rec}.
For testing, similar to \citep{sun2019bert4rec,kang2018selfattentive}, the final interaction of each user is used for test, while the second-to-last interaction is used for validation, with all other interactions forming the training set.

\subsection{Knowledge Base Creation}
\label{appendix:kb}
All our experiment were performed with 5 different values of batch size: $32,64,128,256,512$, and with three different values of learning rate $10^{-3},10^{-4},10^{-5}$. We tried to use values which are commonly used in literature. Lastly, in order to study the influence of the size of the dataset, and consequently the complexity of the task, on the energy consumption and on the test performance of the corresponding training, we removed different percentages of samples from
the data, discarding the same percentage for each of the
classes. In particular, we performed our experiments initially with the entire dataset and then we removed 30\% and 70\% of the samples from the dataset.

Considering all the datasets used for the experiments, we have a total of \nexps experiments, divided into:
\begin{itemize}
    \item 989 computer vision experiments, of which 252 configurations are used for testing;
    \item 637 recommendation systems experiments, of which 180 configurations are used for testing;
    \item 141 natural language processing experiments, of which 36 configurations are used for testing;
\end{itemize}

\subsection{Features extraction}

All the features available in \dname can be found in \cref{tab:features}. Some of them are the same for all the tasks, while others are task-specific. The task features are extracted using Python code computing data statistics. The infrastructural features are extracted using CodeCarbon\footnote{\url{https://codecarbon.io}} library, which allows to extract hardware-specific information. The FLOPS and the number of parameters of the models are extracted using DeepSpeed\footnote{\url{https://www.deepspeed.ai/}} library. All the other model features are extracted using the information available using Pytorch \footnote{\url{https://pytorch.org/}} library, except for LoRA rank in Attention Layers, extracted using HuggingFace\footnote{\url{https://huggingface.co/}} library and Python code, as the mean sequence length, maximum sequence length, Mean Flesch–Kincaid Grade level \citep{kincaid1975derivation} and Mean Dale-Chall Readability score \citep{dale1948formula}.
All the recommendation features are extracted using EasyRec library \citep{betello2024reproducible}.

In order to deal only with numerical features, the few textual features (i.e., the type of activation function) are binarized. Regarding samples with different length, we use padding. This is because there could be models with 6 batch normalization layers, each with its own characteristics, while other models can have 10 batch normalization layers. We used a padding value which our network is able to recognize and the padding length is equal to the length of the longest list.

\begin{table}[!h]
    \centering
    \resizebox{\textwidth}{!}{%
    \begin{tabular}{c|c|c}
        \toprule
        \textbf{Metric} & \textbf{Similarities} & \textbf{Differences} \\
        \midrule
        \textbf{NDCG} & 
        \begin{tabular}{@{}c@{}} Uses rank-based weighting \\ (higher-ranked items matter more). \end{tabular} & 
        \begin{tabular}{@{}c@{}} NDCG is relevance-based and does not \\ consider multiple objectives or absolute \\ differences in values. It evaluates a single ranking. \end{tabular} \\
        \midrule
        \textbf{Kendall’s Tau} & 
        \begin{tabular}{@{}c@{}} Measures ranking consistency \\ between two sets. \end{tabular} & 
        \begin{tabular}{@{}c@{}} SOVA@k incorporates true values in ranking, \\ rather than just rank positions. \end{tabular} \\
        \midrule
        \textbf{Spearman’s Rank Correlation} & 
        \begin{tabular}{@{}c@{}} Measures monotonic relationships \\ between rankings. \end{tabular} & 
        \begin{tabular}{@{}c@{}} Spearman’s method is a purely ordinal measure \\ and does not use value-based distance like SOVA@k. \end{tabular} \\
        \midrule
        \textbf{Hausdorff Distance} & 
        \begin{tabular}{@{}c@{}} Measures the largest distance \\ between points in two sets. \end{tabular} & 
        \begin{tabular}{@{}c@{}} Hausdorff applies in geometric spaces, \\ while SOVA@k operates on ranked sets \\ of multi-objective scores. \end{tabular} \\
        \midrule
        \textbf{$\Delta HV$} & 
        \begin{tabular}{@{}c@{}} Compares two Pareto fronts \\ based on dominated space. \end{tabular} & 
        \begin{tabular}{@{}c@{}} $\Delta HV$ focuses on set coverage, \\ while SOVA@k compares rankings at a fixed $k$. \end{tabular} \\
        \midrule
        \textbf{Borda Count} & 
        \begin{tabular}{@{}c@{}} Uses weighted scores for decision-making \\ across multiple criteria. \end{tabular} & 
        \begin{tabular}{@{}c@{}} SOVA@k does not aggregate rankings but \\ measures distance from an ideal ranking. \end{tabular} \\
        \bottomrule
    \end{tabular}%
    }
    \caption{Comparison of SOVA@k with Existing Metrics}
    \label{tab:metric_comparison}
\end{table}

\section{Description of  Set-Based Order Value Alignment (SOVA) metric} \label{app:sova_metric}

\paragraph{Boundedness of SOVA@k} 

To ensure that the Set-Based Order Value Alignment at k (SOVA@k) is well-defined and interpretable, we prove that it is always bounded within the interval \([0,1]\). 

\begin{lemma} \label{lemma:lower}
    $$\min_{X,Y}\mathrm{SOVA(X,Y)@k} = 0 $$
\end{lemma}
\begin{proof}
    The SOVA@k metric is a sum of non-negative terms:
    \[
    \mathrm{SOVA}(X, Y)@k = \sum_{i=1}^{k} w_i \cdot \sum_{j=1}^{m} \tau_j' \cdot |x_{i,j} - y_{i,j}|,
    \]
    where \(w_i, \tau_j' \geq 0\) (by construction) and \(|x_{i,j} - y_{i,j}| \geq 0\) (since absolute differences are non-negative). Thus, \(\mathrm{SOVA}(X, Y)@k \geq 0\).

    To show attainability, let \(X = Y\). Then \(x_{i,j} = y_{i,j}\) for all \(i, j\), so \(|x_{i,j} - y_{i,j}| = 0\). Substituting into SOVA@k:
    \[
    \mathrm{SOVA}(X, X)@k = \sum_{i=1}^{k} w_i \cdot \sum_{j=1}^{m} \tau_j' \cdot 0 = 0.
    \]
    Therefore, the minimum value of SOVA@k is achievable and equals 0.
\end{proof}

\begin{lemma} \label{lemma:upper}
    $$\max_{X,Y}\mathrm{SOVA(X,Y)@k} = 1 $$
\end{lemma}
\begin{proof}
    Since objectives are normalized to \([0, 1]\), we have \(|x_{i,j} - y_{i,j}| \leq 1\) for all \(i, j\). Substituting into the metric:
    \[
    \mathrm{SOVA}(X, Y)@k = \sum_{i=1}^k w_i \sum_{j=1}^m \tau_j' \cdot |x_{i,j} - y_{i,j}| \leq \sum_{i=1}^k w_i \sum_{j=1}^m \tau_j' \cdot 1 = \left(\sum_{i=1}^k w_i\right)\left(\sum_{j=1}^m \tau_j'\right) = 1.
    \]
    To show attainability, suppose there exist Pareto frontiers \(X\) and \(Y\) where \(x_{i,j} = 1\) and \(y_{i,j} = 0\) for all \(i, j\). This satisfies \(|x_{i,j} - y_{i,j}| = 1\). Substituting into SOVA@k:
    \[
    \mathrm{SOVA}(X, Y)@k = \sum_{i=1}^k w_i \sum_{j=1}^m \tau_j' \cdot 1 = 1.
    \]
    Thus, the maximum value of SOVA@k is 1.
\end{proof}

In this section, we have proved that SOVA@k is mathematically bounded between 0 and 1, ensuring that it remains a well-defined and interpretable ranking alignment metric. The weighting mechanisms—rank-based decay and objective weighting—preserve this property while allowing flexibility in prioritizing objectives and rank positions.

\subsection{Expanded definition of SOVA@K with potential ties in ranks} \label{app:sova_expanded}
The core idea is that if several points in $Y$ (in our application, the predicted Pareto front) share rank \(i\), we treat them as a single “group” for that rank and average their objective-wise differences against the corresponding \(\,x_{i}^{(1)}\,\) in $X$ (in our case, the true Pareto front).

\begin{definition}[SOVA@k with Ties in Ranks]
Let \(X = (x_1, \dots, x_k)\) and \(Y = (y_1, \dots, y_k)\) be two ranked Pareto frontiers of size \(k\), where \(x_i \in I\) is the item at rank \(i\) in \(X\) (the true Pareto front), \(y_i \subseteq I\) is the set of items assigned to rank \(i\) in \(Y\) (the predicted Pareto front). With $\Gamma$ we define the set of items that are ranked at the same position. All objectives \(x_{i,j}\) and \(y_{p,j}\) (for \(x_i \in X\), \(y_p \in \Gamma\)) are normalized to \([0, 1]\).
Let \(w_i\) (position weights) and \(\tau_j'\) (objective weights) satisfy \(\sum_{i=1}^k w_i = 1\) and \(\sum_{j=1}^m \tau_j' = 1\). The \textbf{Set-Based Order Value Alignment at \(k\)} is defined as:
\[
  \mathrm{SOVA}(X, Y)@k 
  \;=\; 
  \sum_{i=1}^{k} 
    w_i 
    \;\cdot\; 
    \sum_{j=1}^{m} 
      \tau_j' 
      \;\cdot\; 
      \Biggl(\frac{1}{|\Gamma_i|}
         \sum_{p \in \Gamma_i}
           \bigl|\,x_{i,j} - y_{p,j}\bigr|
      \Biggr),
\]
where \(x_{i,j}\) and \(y_{p,j}\) denote the \(j\)-th objective value of the \(i\)-th item in \(X\) and the \(p\)-th item in \(\Gamma_i\), respectively.
\end{definition}

\noindent
\paragraph{Boundedness of SOVA@k with Ties}  
Under the assumptions of normalized objectives and normalized weights, \(\mathrm{SOVA}(X, Y)@k \in [0, 1]\).

\begin{proof}  
\textbf{Non-negativity (\( \mathrm{SOVA}(X, Y)@k \geq 0 \)):}  
Since \( |x_{i,j} - y_{p,j}| \geq 0 \), \( w_i > 0 \), and \( \tau_j' \geq 0 \), every term in the summation is non-negative. Thus, \(\mathrm{SOVA}(X, Y)@k \geq 0\).

\textbf{Upper bound (\( \mathrm{SOVA}(X, Y)@k \leq 1 \)):}  
\begin{enumerate}
    \item \textbf{Per-objective difference bound:}  
    Since objectives are normalized, \( |x_{i,j} - y_{p,j}| \leq 1 \) for all \(i, j, p\).  

    \item \textbf{Averaging within a rank:}  
    For rank \(i\), we might have multiple points in \(\Gamma_i\). Taking an average of numbers in \([0,1]\) cannot exceed \(1\). Thus:
    \[
      \frac{1}{|\Gamma_i|}
      \sum_{p \,\in\, \Gamma_i}
        \bigl|\,x_{ij}^{(1)} - x_{pj}^{(2)}\bigr|
      \;\;\in\;\;[0,1].
    \]

    \item \textbf{Weighted summation over objectives:}  
    Since \( \sum_{j=1}^m \tau_j' = 1 \), we have:
    \[
    \sum_{j=1}^m \tau_j' \cdot \Biggl(\frac{1}{|Y_i|} \sum_{y_p \in Y_i} |x_{i,j} - y_{p,j}|\Biggr) \leq \sum_{j=1}^m \tau_j' \cdot 1 = 1.
    \]

    \item \textbf{Weighted summation over ranks:}  
    Since \( \sum_{i=1}^k w_i = 1 \), the final metric satisfies:
    \[
    \mathrm{SOVA}(X, Y)@k \leq \sum_{i=1}^k w_i \cdot 1 = 1.
    \]
\end{enumerate}

\textbf{Attainability of bounds:}  
\begin{itemize}
    \item \textbf{Lower bound (0):} Achieved when \(X = Y\) (i.e., \(Y_i = \{x_i\}\) for all \(i\)), making \( |x_{i,j} - y_{p,j}| = 0 \).  
    \item \textbf{Upper bound (1):} Achieved if \(x_{i,j} = 1\) and \(y_{p,j} = 0\) (or vice versa) for all \(i, j, p\), under valid Pareto dominance.  
\end{itemize}

Thus, \( 0 \leq \mathrm{SOVA}(X, Y)@k \leq 1 \).  
\end{proof}

\paragraph{Key Features of the Metric}

\begin{itemize}

\item \textbf{Rank-Based Weighting}: Each ranking position is assigned a weight that decreases exponentially as the rank increases, prioritizing the accuracy of higher-ranked points. As a consequence, errors in higher-ranked points are treated as more significant.
\item \textbf{Objective Weighting}: Each objective is assigned a relative weight, allowing users to prioritize specific objectives.The user-provided weights are normalized directly to sum to 1.

\item \textbf{Distance Aggregation}: The absolute differences between corresponding ranking positions in the two sets are calculated and weighted by their rank and objective importance. The total weighted distance is then aggregated across all ranks.

\item \textbf{[0-1] Bound}:The metric is bounded in the range [0,1]:  a value of 0 indicates perfect alignment between the rankings of the two sets, while value of 1 represents the maximum possible disagreement.

\end{itemize}

\section{Experimental setting}
\label{app:exp_set}
\subsection{Hardware Specification}
\label{app:hardware_spec}
We have used three different gpu across our experiments:
\begin{itemize}
    \item NVIDIA A100-SXM with 80GB of VRAM, equipped with a AMD EPYC 7742 64-Core Processor cpu.
    \item NVIDIA GeForce RTX 4090 with 24GB of VRAM, equipped with a AMD Ryzen 9 7900 12-Core Processor cpu.
    \item NVIDIA L40S with 45GB of VRAM, equipped with a AMD EPYC 7R13 Processor cpu.
\end{itemize}

\subsection{Standard metrics to assess quality of Pareto solutions}
\label{app:metrics}

\paragraph{Hausdorff distance}
The Hausdorff distance quantifies the maximum deviation between the predicted and true Pareto fronts:
\begin{equation*}   
d_H(\mathcal{P}_{\text{pred}}, \mathcal{P}_{\text{true}}) = 
\max\Bigg\{
\sup_{p \in \mathcal{P}_{\text{pred}}} \inf_{q \in \mathcal{P}_{\text{true}}} \|p - q\|,
\sup_{q \in \mathcal{P}_{\text{true}}} \inf_{p \in \mathcal{P}_{\text{pred}}} \|p - q\|
\Bigg\}.
\end{equation*}

where $\|p - q\|$ is the Euclidean distance between solutions $p$ and $q$ in the objective space. Since in our scenario, the objective are bounded in $[0, 1]$, the a Hausdorff distance range of $[0, \sqrt{2}]$. Smaller values of $d_H$ indicate better alignment between the predicted and true fronts.
\paragraph{Hypervolume}
The Hypervolume (HV) quantifies the volume of the region in the objective space that is weakly dominated by a set of solutions and bounded with respect to a given reference point. Given a solution set \( \mathcal{P} \subset \mathbb{R}^m \) and a reference point \( r \in \mathbb{R}^m \), the HV is defined as:
\begin{equation*}
HV(\mathcal{P}) = \lambda \left( \bigcup_{p \in \mathcal{P}} [p, r] \right),
\end{equation*}
where \( [p, r] \) denotes the hyperrectangle formed between point \( p \) and the reference point \( r \), and \( \lambda \) is the Lebesgue measure in \( \mathbb{R}^m \), representing the volume. A larger HV value indicates that a greater portion of the objective space is dominated by \( \mathcal{P} \), implying a better approximation to the true front. When comparing two fronts, we compute the \textit{Hypervolume difference}:
\begin{equation*}
\Delta HV = HV(\mathcal{P}_{\text{true}}) - HV(\mathcal{P}_{\text{pred}}),
\end{equation*}
which captures the volume of the objective space that is dominated by the true front but not by the predicted one. Since the objectives are normalized in \([0, 1]^m\), the HV values are bounded within \([0, 1]\) for \( m = 2 \), and smaller values of \( \Delta HV \) indicate better alignment between the predicted and true Pareto fronts.
\paragraph{NDCG}
This metric evaluates the alignment of solution rankings in the predicted and true Pareto fronts, incorporating weights $(\omega_A, \omega_E)$ to reflect the relative importance of validation performance and energy consumption, respectively, where $\omega_A + \omega_E = 1$. The NDCG at rank $k$ is computed as:
\[
\text{NDCG@}k = \frac{\sum_{i=1}^k \frac{2^{\text{rel}_i} - 1}{\log_2(i + 1)}}{\sum_{i=1}^k \frac{2^{\text{rel}_i^{\text{ideal}}} - 1}{\log_2(i + 1)}},
\]
where $\text{rel}_i$ and $\text{rel}_i^{\text{ideal}}$ denote the relevance scores of the predicted and ideal rankings, respectively. By varying the weights $(\omega_A, \omega_E)$, we analyze ranking consistency under different prioritization preferences.

\paragraph{Recall}
Recall measures the proportion of true Pareto-optimal solutions that are successfully identified in the predicted front. Let \( \mathcal{P}_{\text{true}} \) denote the set of true Pareto-optimal solutions and \( \mathcal{P}_{\text{pred}} \) the predicted ones. Then recall is computed as:
\begin{equation*}
\text{Recall} = \frac{|\mathcal{P}_{\text{pred}} \cap \mathcal{P}_{\text{true}}|}{|\mathcal{P}_{\text{true}}|}.
\end{equation*}
Recall values lie in the range \([0, 1]\), where higher values indicate that a larger fraction of the true Pareto front has been correctly predicted.

\paragraph{F1-score}
The F1-score is the harmonic mean of precision and recall, providing a single measure that balances both aspects. It is particularly useful when one seeks a trade-off between including many relevant solutions (recall) and minimizing false positives (precision). Given precision \( P \) and recall \( R \), the F1-score is defined as:
\begin{equation*}
\text{F1} = 2 \cdot \frac{P \cdot R}{P + R}.
\end{equation*}

\section{Competitors Details}
\label{app:comp_det}
\paragraph{ECNAS}
We use NasBench-101, which restricts the architecture search space to $3\times3$ convolutions, $1\times1$ convolutions, and $3\times3$ max pooling. The original paper reports 10 trials with a population size of 10 over 100 evolutions. We adopt the SEMOA algorithm, identified as the best-performing method in their work. Their code produces a Pareto frontier of DAG-based architectures for each trial. From this, we select two architectures - one that maximises accuracy and another that balances accuracy with 50\% energy consumption. These DAGs are then converted into architectures according to the specifications in the paper and the NasBench-101 GitHub repository\footnote{\url{https://github.com/google-research/nasbench}}. Since several architectures achieved optimal accuracy and balanced emissions, we randomly selected one from the two category considered and test it on CIFAR-10, following the specifications of the original paper, with the epoch budget obtained from the search.

\paragraph{CENAS}
Employs reinforcement learning to optimize NAS algorithms based on GPU availability but is restricted to a narrow set of layer types, e.g. zeroization, skip-connection, 1x1 convolution, 3x3 convolution, and 3x3 average pooling. While CENAS is expected to return a Pareto frontier similar to EC-NAS, the available code does not output the architectures, making it difficult to analyze or reproduce results. Attempts to contact the authors for clarification went unanswered, leaving key implementation details uncertain.

\paragraph{KNAS}
This approach prioritizes efficiency and sustainability during the architecture search process but does not account for emissions from the final trained model. It uses the same layer types of CENAS. Additionally, unlike other NAS methods, it does not produce a Pareto frontier, making it less transparent in terms of trade-offs between accuracy, efficiency, and resource consumption. Despite these limitations, we selected a model based on its reported performance and trained it using the original paper’s specifications to ensure a fair comparison. As well as EC-NAS, we test the selected architecture using CIFAR-10.

\section{Technical Addendum}
\label{app:additional_studies}

\subsection{Pareto Front Extraction and Filtering}
\begin{figure}[t!]
    \centering
    \includegraphics[width=0.7\textwidth]{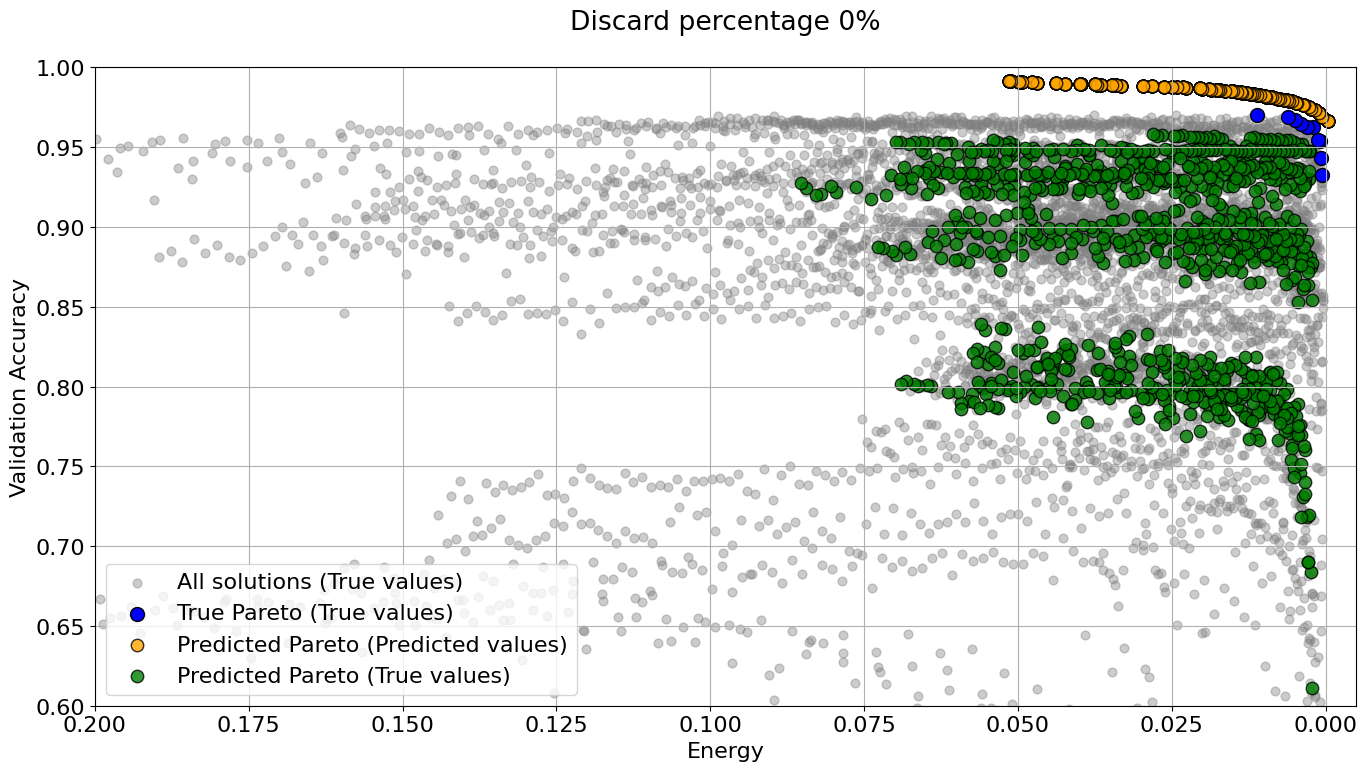}
    \vspace{0.5em} 
    \includegraphics[width=0.7\textwidth]{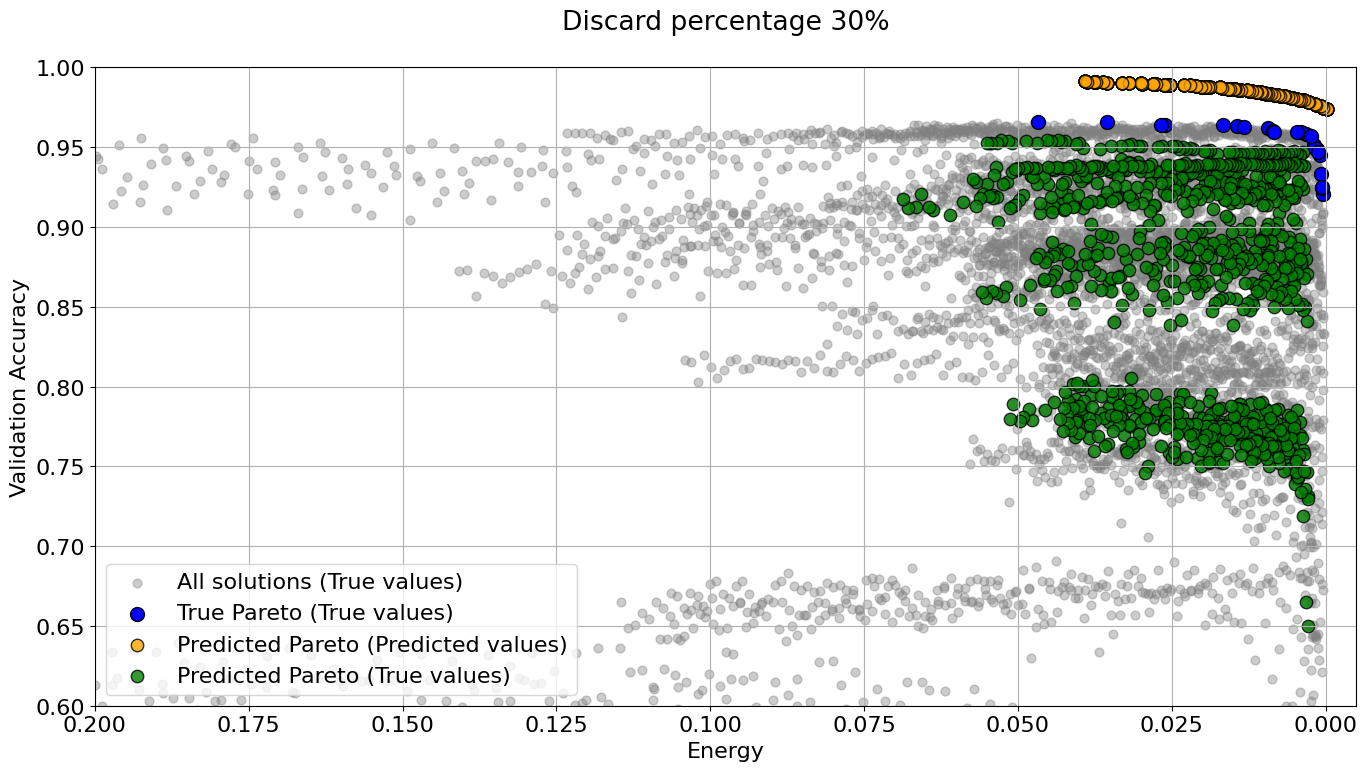}
    \vspace{0.5em}
    \includegraphics[width=0.7\textwidth]{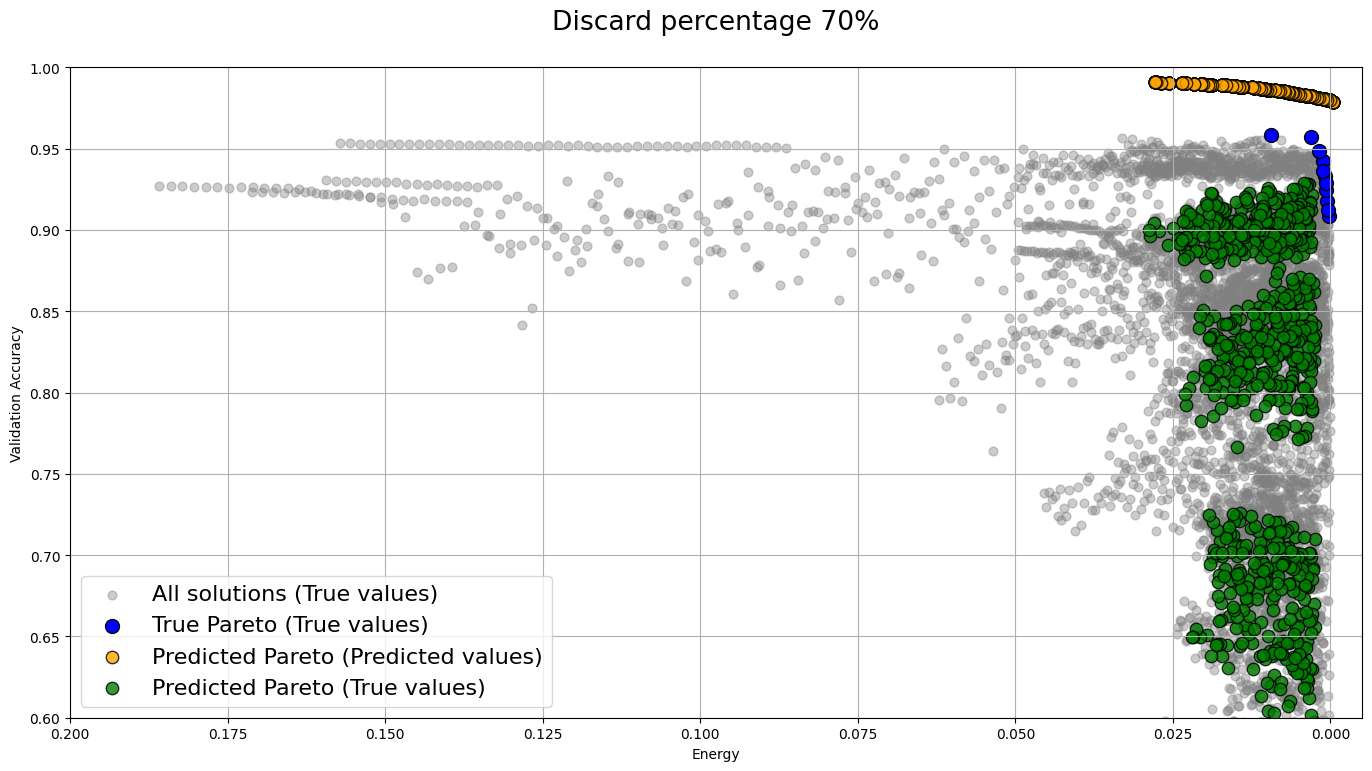}
    \caption{
        \textbf{Comparison of True and Predicted Pareto Fronts on CIFAR-10}.
        Pareto-optimal configurations are shown in the normalized Validation Accuracy vs. Energy space. Each subfigure corresponds to a different percentage of discarded test data (0\%, 30\%, 70\%), while the predictor is trained with the same seed (42) in all cases. Gray dots represent all configurations evaluated with true target values. Blue markers show the \textit{True Pareto front}, orange markers the \textit{Predicted Pareto front} based on the predicted value of the objectives and green markers denote \textit{Predicted Pareto configurations re-evaluated with true values}.
        Both true and predicted fronts include only configurations achieving at least \textbf{0.9 validation accuracy}, filtered based on their respective values.
        The \textit{x}-axis (Energy) is limited to the normalized range [0.00, 0.20], and the \textit{y}-axis (Validation Accuracy) spans [0.6, 1.0] to enhance clarity.}
    \label{fig:cifar10_pareto_vertical}
\end{figure}

\begin{figure}[t!]
    \centering
    \includegraphics[width=0.7\textwidth]{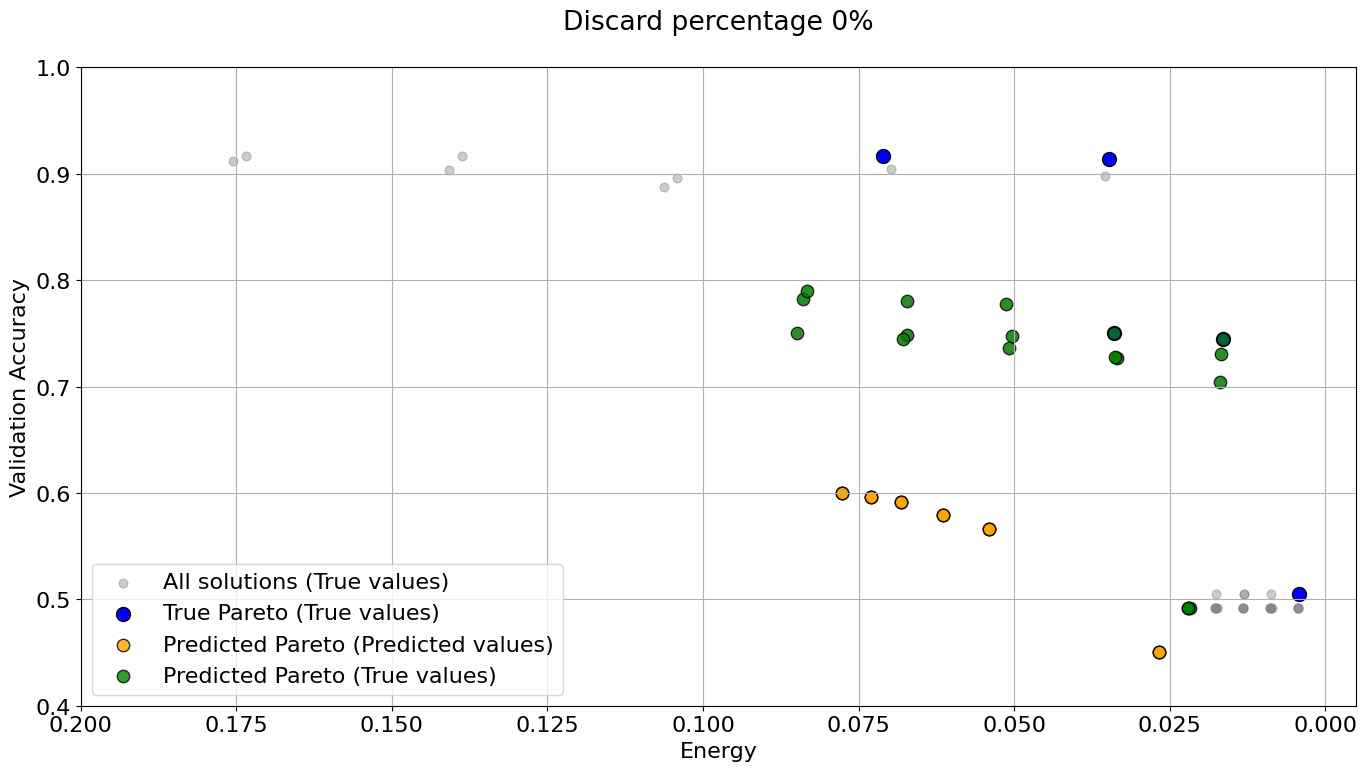}
    \vspace{0.5em} 
    \includegraphics[width=0.7\textwidth]{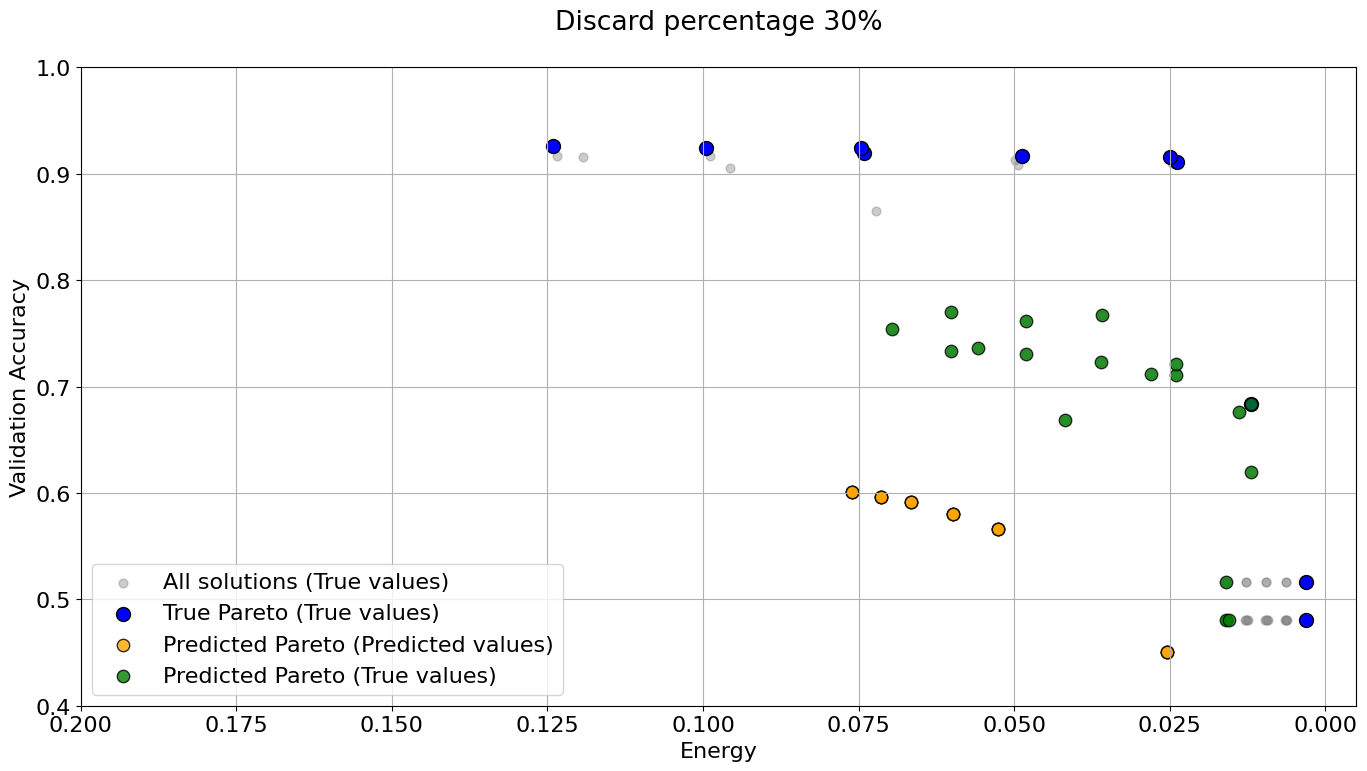}
    \vspace{0.5em}
    \includegraphics[width=0.7\textwidth]{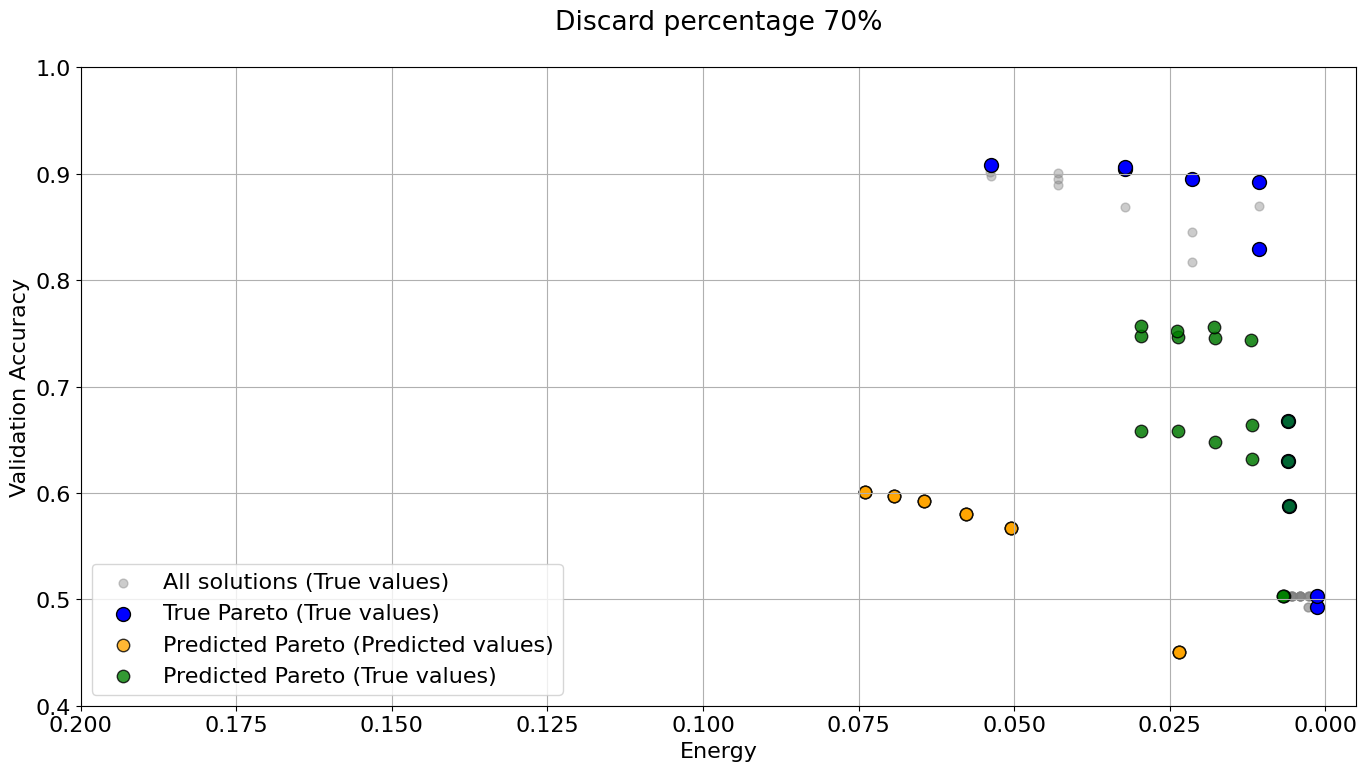}
    \caption{
        \textbf{Comparison of True and Predicted Pareto Fronts on Rotten\_tomatoes}.
        Pareto-optimal configurations are shown in the normalized Validation Accuracy vs. Energy space. Each subfigure corresponds to a different percentage of discarded test data (0\%, 30\%, 70\%), while the predictor is trained with the same seed (42) in all cases.
        Gray dots represent all configurations evaluated with true target values. Blue markers show the \textit{True Pareto front}, orange markers the \textit{Predicted Pareto front} based on the predicted value of the objectives and green markers denote \textit{Predicted Pareto configurations re-evaluated with true values}.
        Both true and predicted fronts include only configurations achieving at least \textbf{0.45 validation accuracy}, filtered based on their respective values.
        The \textit{x}-axis (Energy) is limited to the normalized range [0.00, 0.20], and the \textit{y}-axis (Validation Accuracy) spans [0.4, 1.0] to enhance clarity.}
        \label{fig:rotten_pareto}
\end{figure}

\begin{figure}[t!]
    \centering
    \includegraphics[width=0.7\textwidth]{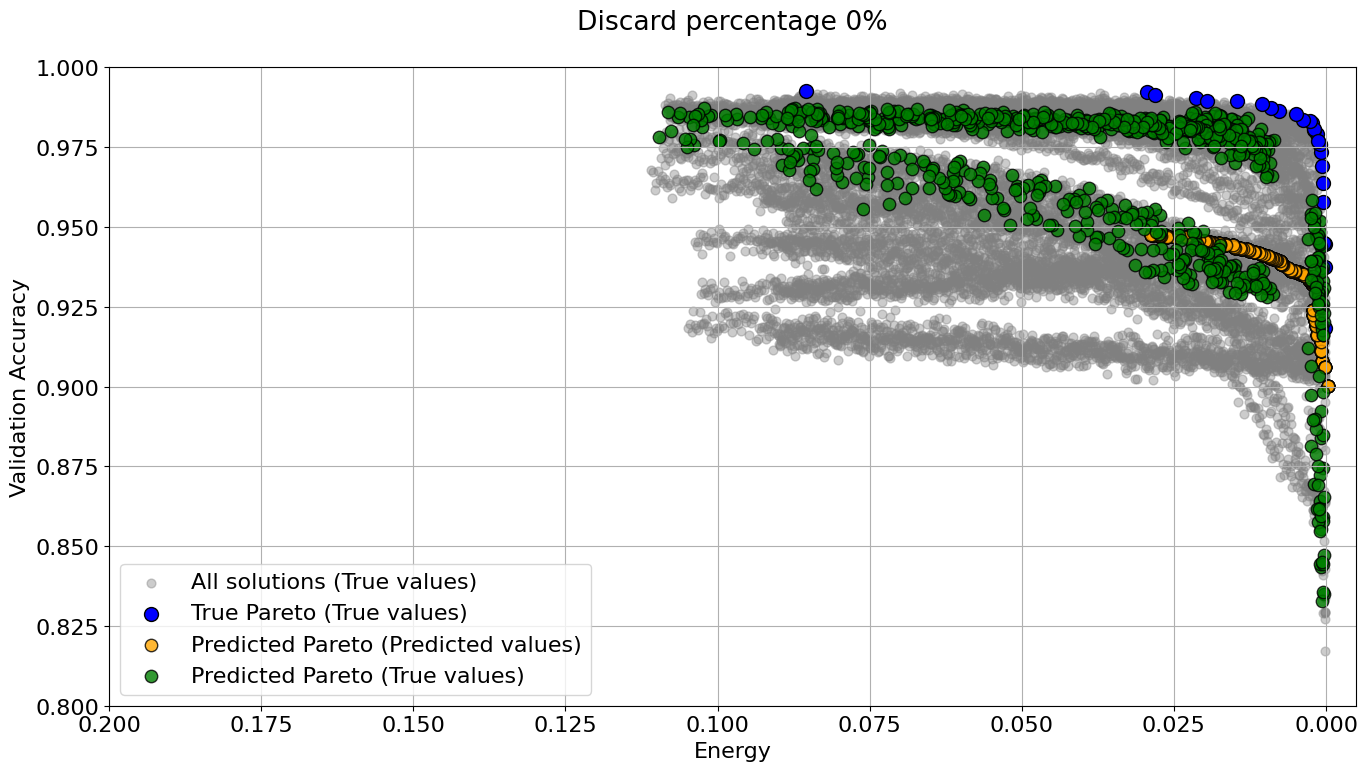} 
    \vspace{0.5em} 
    \includegraphics[width=0.7\textwidth]{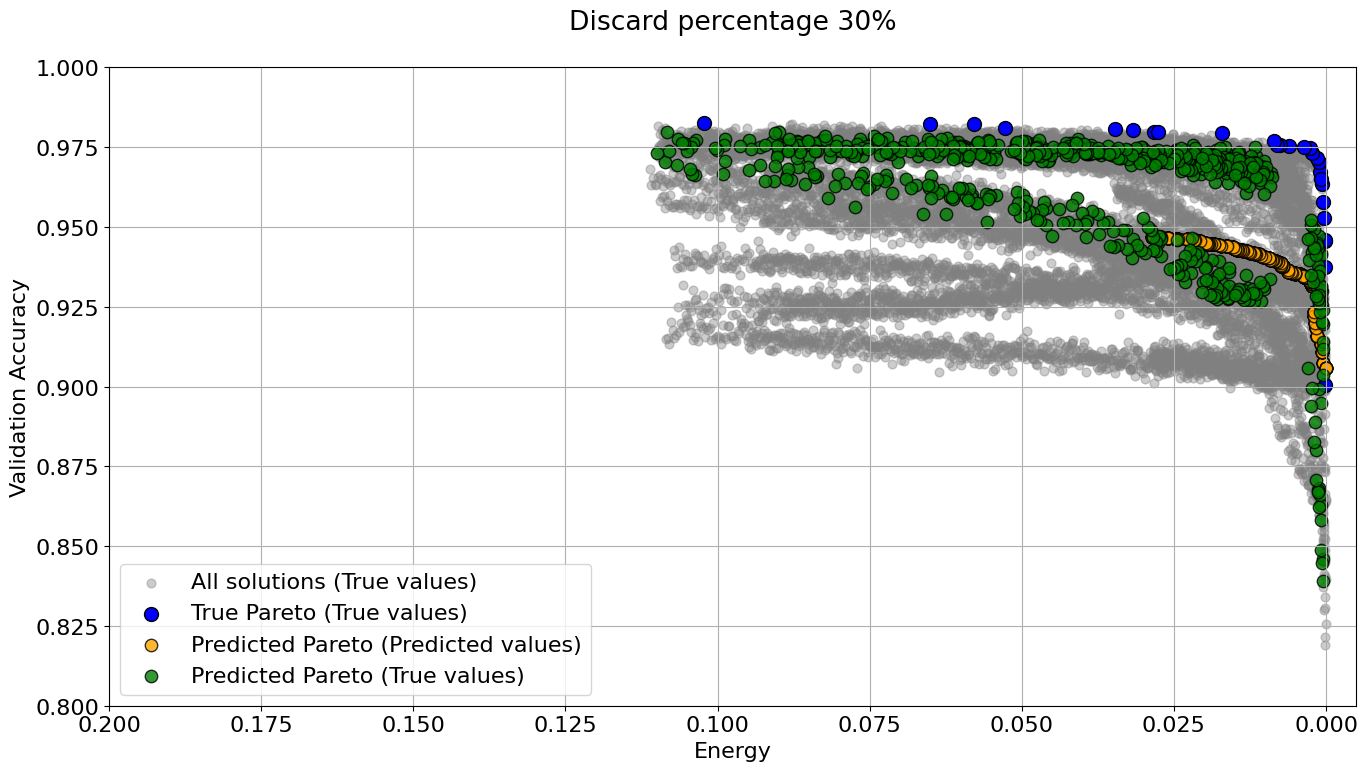}
    \vspace{0.5em} 
    \includegraphics[width=0.7\textwidth]{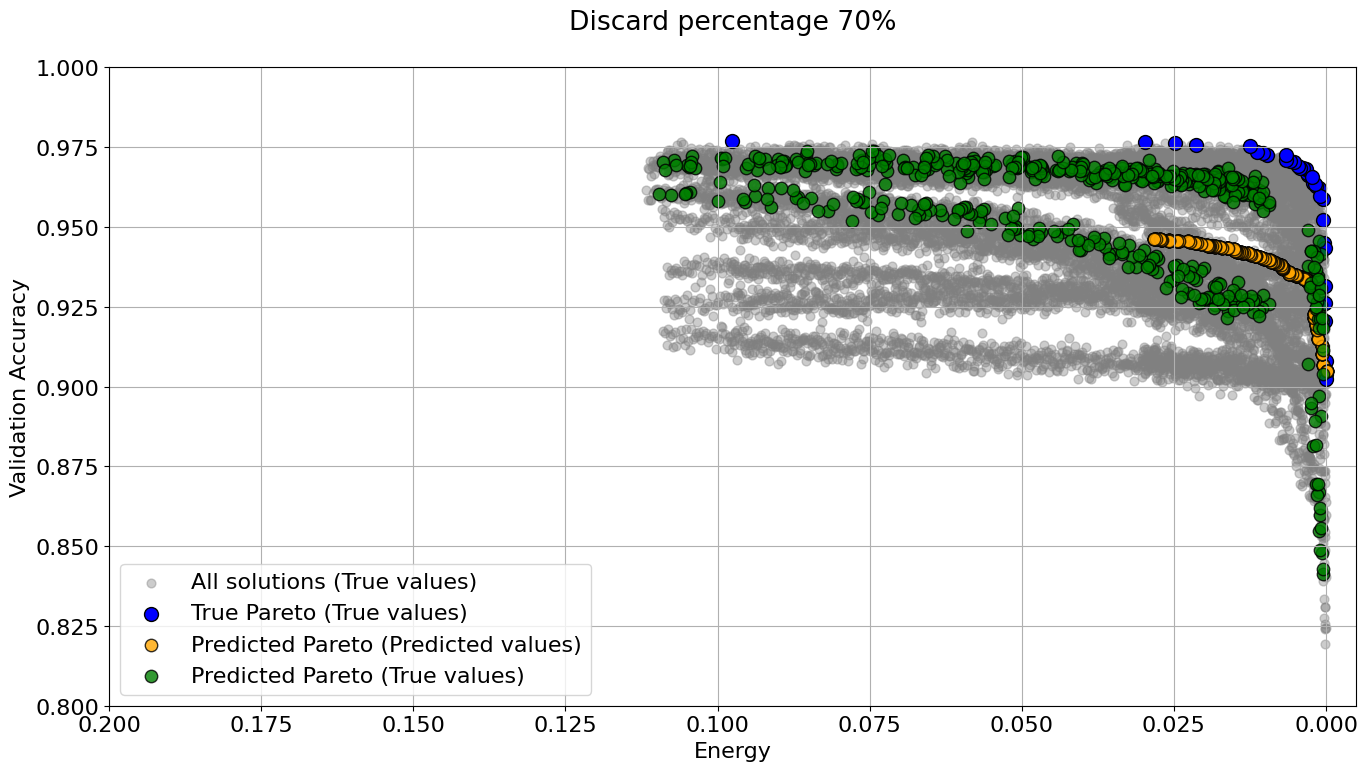} 
    \caption{
        \textbf{Comparison of True and Predicted Pareto Fronts on FS-TKY}.
        Pareto-optimal configurations are shown in the normalized Validation Accuracy vs. Energy space. Each subfigure corresponds to a different percentage of discarded test data (0\%, 30\%, 70\%), while the predictor is trained with the same seed (42) in all cases.
        Gray dots represent all configurations evaluated with true target values. Blue markers show the \textit{True Pareto front}, orange markers the \textit{Predicted Pareto front} based on the predicted value of the objectives and green markers denote \textit{Predicted Pareto configurations re-evaluated with true values}.
        Both true and predicted fronts include only configurations achieving at least \textbf{0.9 validation accuracy}, filtered based on their respective values.
        The \textit{x}-axis (Energy) is limited to the normalized range [0.00, 0.20], and the \textit{y}-axis (Validation Accuracy) spans [0.80, 1.0] to enhance clarity.}
    \label{fig:fsq_pareto_vertical}
\end{figure}

After obtaining predictions for the two objectives at each epoch within the epoch space for each model configuration, we extract the Pareto fronts based on the ground truth and predicted objective metrics. We identify the Pareto-optimal points by checking for non-domination: a point is considered Pareto-optimal if there is no other point that is better in all objectives and strictly better in at least one objective. This process ensures that the selected points form the Pareto front, representing the best trade-offs among the objectives. Once identified the Pareto fronts, we apply a filtering step based on a user-defined threshold for the validation accuracy objective. This filter removes any solutions not meeting the required accuracy, focusing the analysis on the most relevant solutions for the task at hand. Figures \ref{fig:cifar10_pareto_vertical}, \ref{fig:fsq_pareto_vertical}, \ref{fig:rotten_pareto} show the obtained Pareto curves on the three test datasets.

\subsection{Running Time  and Time complexity Analysis}
While traditional NAS algorithms require a complete re-run of the search process for each new dataset, \approach adopts a different approach. By representing both datasets and models through features \approach can operate directly at inference time, eliminating the need for repeated searches.

As for the time complexity of \approach, it is primarily driven by its two main components: the target predictor and the ranker for Pareto solution. As for the first component, the complexity of the standard transformer attention mechanism per layer is $\mathcal{O}(L^2 \cdot H \cdot \frac{E}{H}) = \mathcal{O}(L^2 \cdot E)$ where $H$ is the number of heads of the transformer, $L$ is the sequence length and $\frac{E}{H}$ represents the size of each head. For $A$ attention blocks and batch size $B$, the total complexity becomes $\mathcal{O}(B \cdot L^2 \cdot E \cdot A)$. Here, the quadratic term $L^2$ dominates the computation for long sequences. The time complexity for computing the Pareto front and ranking solutions in our approach is primarily determined by the full Pareto front selection process, which has a time complexity of $\mathcal{O}(m \cdot n^2)$, where n is the number of points in the dataset and m is the number of objectives (2). This step involves checking the dominance of each point against every other point, resulting in quadratic complexity. After the Pareto front is computed, we apply a filtering step based on the minimum accuracy threshold for a specific objective, which has a time complexity of $\mathcal{O}(n)$. This filtering step reduces the number of points considered in the subsequent ranking process. Following filtering, the ranking and normalization operations involve both normalization of the objectives $\mathcal{O}(m \cdot n)$ and weighted scoring with sorting $\mathcal{O}(m \cdot n + n\log n)$, resulting in an overall complexity of $\mathcal{O}(m \cdot n + n\log n)$. Thus, the dominant factor in the overall complexity is the $\mathcal{O}(m \cdot n^2)$ complexity for the Pareto front selection. For our specific application, where the number of points is limited, this approach is well-suited. However, for larger datasets, algorithms with lower computational complexity can be applied.

\subsection{Predictor Sanity Check}
To assess whether the predictive function \( q_{\theta} \) has learned meaningful properties beyond memorization, we performed a sanity check based on input perturbation. Specifically, we compared the original predictions reported in the paper with those obtained by duplicating each training example and halving the number of training epochs, keeping all other conditions fixed. This modification preserves the overall number of gradient updates while altering the training dynamics. The resulting differences in prediction accuracy, measured in terms of MAE for both objectives, are reported in \cref{tab:sanity_check}. The small differences observed between the two configurations testify to the robustness of \( q_{\theta} \), suggesting that the model captures generalizable patterns rather than overfitting to specific training dynamics.

\begin{table}[ht]
\centering
\small
\begin{tabular}{lccc}
\toprule
\textbf{Dataset} & \textbf{Discard percentage (\%)} & $\Delta\text{MAE}_A$ & $\Delta\text{MAE}_E$ \\
\midrule
\multirow{3}{*}{CIFAR-10} & 0 & 0.00474 & 0.00273 \\
& 30 & 0.00629 & 0.00426 \\
&70 & 0.01370 & 0.00760 \\
\midrule
\multirow{3}{*}{Rotten\_tomatoes} & 0  & 0.00519 & 0.00472 \\
& 30 & 0.00005 & 0.00267 \\
& 70 & 0.01110 & 0.00170 \\
\midrule
\multirow{3}{*}{FS-TKY} & 0  & 0.00301 & 0.00274 \\
 & 30 & 0.00251 & 0.00088 \\
& 70 & 0.00229 & 0.00087 \\
\bottomrule
\end{tabular}
\caption{Difference in MAE metrics across test datasets, comparing the predictions presented in the paper with those obtained by duplicating training examples and halving the number of training epochs (seed = 476).}
\label{tab:sanity_check}
\end{table}

\section{Additional Results}
\label{app:additional_results}

In this section, we present additional results from our experiments. \cref{tab:swee_model} shows the performance of \approach on each of the models for the 3 different datasets. The performance remain consistent with the results obtained in \cref{sec:results}, showing that the proposed approach is able to achieve good performance on each of the selected models from \dname.

\cref{tab:sweep_lr} does something similar changing the values of learning rate, evidencing that with lower learning rate values, \approach is able to better predict the performance of the selected network. This is more evident for the accuracy performance, while the predicted energy is always close to the ground truth.

\begin{table}[ht]
    \centering
    \begin{tabular}{c|ccc}
    \toprule
    Dataset & Model & MAE$_A$ ($\downarrow$) & MAE$_E$ ($\downarrow$) \\
    \midrule
\multirow{6}{*}{CIFAR-10} & VIT & 0.181 $\pm$ 0.000 & 0.034 $\pm$ 0.004 \\
 & AlexNet & 0.203 $\pm$ 0.008 & 0.004 $\pm$ 0.000 \\
 & SqueezeNet & 0.099 $\pm$ 0.021 & 0.006 $\pm$ 0.002 \\
 & ResNet18 & 0.079 $\pm$ 0.010 & 0.007 $\pm$ 0.000 \\
 & EfficientNet & 0.041 $\pm$ 0.005 & 0.010 $\pm$ 0.000 \\
 & VGG16 & 0.228 $\pm$ 0.013 & 0.013 $\pm$ 0.003 \\
\midrule
\multirow{4}{*}{Rotten\_tomatoes} & RoBERTa & 0.063 $\pm$ 0.035 & 0.007 $\pm$ 0.002 \\ 
 & BERT & 0.045 $\pm$ 0.014 & 0.011 $\pm$ 0.002 \\
 & Mistral-7B-v0.3 & 0.314 $\pm$ 0.023 & 0.053 $\pm$ 0.006 \\
 & Microsoft-PHI-2 & 0.131 $\pm$ 0.016 & 0.027 $\pm$ 0.003 \\ 
\midrule
\multirow{4}{*}{FS-TKY} & SASRec & 0.026 $\pm$ 0.002 & 0.030 $\pm$ 0.001 \\ 
& GRU4Rec & 0.022 $\pm$ 0.001 & 0.035 $\pm$ 0.001 \\
& BERT4Rec & 0.026 $\pm$ 0.002 & 0.029 $\pm$ 0.001 \\
& CORE & 0.020 $\pm$ 0.003 & 0.030 $\pm$ 0.001 \\
\bottomrule
    \end{tabular}
    \caption{MAE of the predicted performance (A for accuracy and E for energy) with respect to the ground truth obtained from \dname. This table shows the performance of \approach on each of the models for the 3 different datasets. Overall, for each task, the performance remain consistent across the models. Some outliers (i.e., Mistral-7B-v0.3) could depend on the reduced number of epochs that we selected to train the models.}
    \label{tab:swee_model}
\end{table}

\begin{table}[h]
    \centering
    \begin{tabular}{c|ccc}
    \toprule
    Dataset & Learning Rate & MAE$_A$ ($\downarrow$) & MAE$_E$ ($\downarrow$) \\
      \midrule 
      \multirow{2}{*}{CIFAR-10} &  $10^{-3}$ & 0.043 $\pm$ 0.012 & 0.012 $\pm$ 0.001 \\ 
      & $10^{-2}$ & 0.225 $\pm$ 0.022 & 0.011 $\pm$ 0.002 \\
      \midrule
      \multirow{2}{*}{FS-TKY} & $10^{-3}$ & 0.013 $\pm$ 0.002 & 0.031 $\pm$ 0.001 \\
      & $10^{-2}$ &0.034 $\pm$ 0.002 & 0.031 $\pm$ 0.001 \\
      \midrule
      \multirow{2}{*}{Rotten\_tomatoes} & $10^{-3}$ & 0.115 $\pm$ 0.021 & 0.022 $\pm$ 0.002 \\
      & $10^{-2}$ &  0.145 $\pm$ 0.021 & 0.024 $\pm$ 0.002 \\
      \bottomrule
    \end{tabular}
    \caption{Mean Absolute Error of the predicted performance (A for accuracy and E for energy) with respect to the ground truth obtained from \dname. This table shows the performance of \approach on two different learning rate values selected in our test set. This plots makes evident that with lower learning rate values, \approach is able to better predict the performance of the selected network.}
    \label{tab:sweep_lr}
\end{table}

\begin{table*}[!t]
  \centering
  \begin{tabular}{c|l}
    \toprule
    Architectural component \\
    \midrule
    Batch Size & \multirow{2}{*}{$\vphantom{\begin{array}{c}X\\X\\\end{array}}$
    \begin{tabular}[c]{@{}l@{}}
    Hyperparameters \end{tabular}} \\ 
    Learning rate \\
    \midrule
    Number of classes \custombox[black]{CV} \custombox[mydarkgreen]{NLP}  & \multirow{2}{*}{$\vphantom{\begin{array}{c}X\\X\\\end{array}}$
    \begin{tabular}[c]{@{}l@{}} Task \\ features \end{tabular}} \\ Class distribution \custombox[black]{CV} \custombox[mydarkgreen]{NLP} \\ 
    \midrule
    GPU L2 cache size  & \multirow{8}{*}{$\vphantom{\begin{array}{c}X\\X\\X\\X\\X\\X\\X\\X\\\end{array}}$
    \begin{tabular}[c]{@{}l@{}} Infrastuctural \\ features \end{tabular}} \\ GPU major revision number \\
    GPU minor revision number \\
    GPU total memory \\
    GPU multi processor count \\
    CPU bits \\
    CPU number of cores \\
    CPU Hz advertised \\   
    \midrule
    FLOPS & \multirow{24}{*}{$\vphantom{\begin{array}{c}X\\X\\X\\X\\X\\X\\X\\X\\X\\X\\X\\X\\X\\X\\X\\X\\X\\X\\X\\X\\X\\X\\X\\X\\\end{array}}$
    \begin{tabular}[c]{@{}l@{}} Model \\ features \end{tabular}} \\
    Number of parameters \\
    Total number of layers  \\
    Type of activation functions \\
    Number of Convolutional layers \custombox[black]{CV} \\
    Dimension of Output Channels of Convolutional Layers \custombox[black]{CV} \\
    Kernel Size of Convolutional Layers \custombox[black]{CV} \\
    Stride of Convolutional Layers \custombox[black]{CV} \\
    Number of Fully Connected Layers \\
    Input Features of Fully Connected Layers \\
    Number of Attention layers \\
    Type of Attention Layers \\
    Input Features of Attention Layers \\
    Number of Heads in Attention Layers \\
    LoRA rank in Attention Layers \custombox[mydarkgreen]{NLP} \\
    Number of Embedding Layers \\
    Embedding Dimension of Embedding Layers \\
    Number of Batch Normalization Layers \\
    Numerical Stability $\epsilon$ of Batch Normalization Layers \\
    Momentum of Batch Normalization Layers  \\
    Number of Layer Normalization Layers \\
    Numerical Stability $\epsilon$ of Layer Normalization Layers \\
    Number of Dropout Layers \\
    Dropout Probability of Dropout Layers \\
    \midrule
    Discard percentage  & \multirow{16}{*}{$\vphantom{\begin{array}{c}X\\X\\X\\X\\X\\X\\X\\X\\X\\X\\X\\X\\X\\X\\X\\X\\\end{array}}$
    \begin{tabular}[c]{@{}l@{}} Data \\ features \end{tabular}} \\ Number of training examples \custombox[black]{CV} \custombox[mydarkgreen]{NLP} \\
    Number of validation examples \custombox[black]{CV} \custombox[mydarkgreen]{NLP} \\
    Image shape \custombox[black]{CV}\\
    Mean Pixel Values  \custombox[black]{CV}\\
    Pixel Standard Deviation \custombox[black]{CV}\\
    Number of users \custombox[blue]{RS}\\
    Number of items \custombox[blue]{RS}\\
    Number of interactions \custombox[blue]{RS}\\
    Interaction Density \custombox[blue]{RS}\\
    Average Interaction Length \custombox[blue]{RS}\\
    Median Interaction length \custombox[blue]{RS}\\
    Mean sequence length \custombox[mydarkgreen]{NLP}\\
    Maximum sequence length \custombox[mydarkgreen]{NLP}\\
    Mean Flesch–Kincaid Grade level \custombox[mydarkgreen]{NLP}\\
    Mean Dale-Chall Readability score \custombox[mydarkgreen]{NLP} \\
    \bottomrule
  \end{tabular}
  \caption{List of all the features used to build \dname. The features denoted with the \custombox[blue]{RS} label are used only for the recommendation tasks, the ones with the \custombox[black]{CV} label are used only for the computer vision tasks and the ones with the \custombox[mydarkgreen]{NLP} are used only for the natural language processing tasks. The remaining are shared between tasks.}
\label{tab:features}
\end{table*}

\end{document}